\documentclass[journal]{IEEEtran}
\usepackage{amsmath,amssymb,amsfonts,color,subfigure}
\usepackage{graphicx}
\usepackage{setspace}
\usepackage{float}
\usepackage{epsfig,epstopdf}
\usepackage{verbatim}
\usepackage[section]{algorithm}
\usepackage{algorithmic}
\usepackage{array}
\usepackage{amsthm}
\usepackage{booktabs}
\usepackage{graphicx}
\usepackage{subfigure}
\usepackage{multirow}

\usepackage{threeparttable}
\usepackage{hyperref}
\usepackage{makecell}
\usepackage{fmtcount}
\usepackage{booktabs}

\usepackage{nicefrac}
\usepackage{xcolor}
\usepackage{xspace}
\usepackage{enumerate}
\usepackage{mathrsfs}



\newcolumntype{L}[1]{>{\raggedright\let\newline\\\arraybackslash\hspace{0pt}}m{#1}}
\newcolumntype{C}[1]{>{\centering\let\newline\\\arraybackslash\hspace{0pt}}m{#1}}
\newcolumntype{R}[1]{>{\raggedleft\let\newline\\\arraybackslash\hspace{0pt}}m{#1}}

\numberwithin{equation}{section}

\theoremstyle{remark}

\newcommand{\be}{\begin{eqnarray}}
\newcommand{\ben}{\begin{eqnarray*}}
\newcommand{\en}{\end{eqnarray}}
\newcommand{\enn}{\end{eqnarray*}}
\newcommand{\bea}{\begin{aligned}}
\newcommand{\ena}{\end{aligned}}
\newcommand{\cR}{\textcolor[rgb]{1.00,0.00,0.00}}
\newcommand{\cG}{\textcolor[rgb]{0.00,0.00,1.00}}
\newcommand{\cB}{\textcolor[rgb]{0.00,1.00,0.00}}

\graphicspath{{figures/}}

\begin{document}
%
\title{A Fast and Compact Saliency Score Regression Network Based on Fully Convolutional Network}
%
%
%

\author{Xuanyang~Xi, Yongkang~Luo, Fengfu~Li, Peng~Wang and Hong~Qiao
		
\thanks{
X. Xi and H. Qiao are with State Key Laboratory of Management and Control for Complex Systems, Institute of Automation, Chinese Academy of Sciences, Beijing 100190, China. They are also with University of Chinese Academy of Sciences, Beijing 100049, China. In additon, Qiao is with CAS Center for Excellence in Brain Science and Intelligence Technology, Shanghai 200031, China.

Y. Luo and P. Wang are with the Research Center of Precision Sensing and Control, Institute of Automation, Chinese Academy of Sciences, Beijing 100190, China. 

F. Li is with LSEC and Institute of Applied Mathematics, AMSS, Chinese Academy of Sciences, Beijing 100190, China.

       }
}

\maketitle

\begin{abstract}
	
Visual saliency detection aims at identifying the most visually distinctive parts in an image, and serves as a pre-processing step for a variety of computer vision and image processing tasks. To this end, the saliency detection procedure must be as fast and compact as possible and optimally processes input images in a real time manner. It is an essential application requirement for the saliency detection task. However, contemporary detection methods often utilize some complicated procedures to pursue feeble improvements on the detection precession, which always take hundreds of milliseconds and make them not easy to be applied practically.  In this paper, we tackle this problem by proposing a fast and compact saliency score regression network which employs fully convolutional network, a special deep convolutional neural network, to estimate the saliency of objects in images. It is an extremely simplified end-to-end deep neural network without any pre-processings and post-processings. When given an image, the network can directly predict a dense full-resolution saliency map (image-to-image prediction). It works like a compact pipeline which effectively simplifies the detection procedure. Our method is evaluated on six public datasets, and experimental results show that it can achieve comparable or better precision performance than the state-of-the-art methods while get a significant improvement in detection speed (35 FPS, processing in real time).

\end{abstract}

\begin{IEEEkeywords}
Salient object detection, saliency score regression, deep convolutional neural networks, fully convolutional networks, real time.
	
\end{IEEEkeywords}

\section{Introduction}
%
%
%
%
%

Detecting salient attention-grabbing objects and segmenting entire objects from images or videos, without any prior knowledge about the scenes, is the aim of salient object detection \cite{SalObjBenchmark}. It is always considered as a pre-processing step in different tasks, which provides wide applications in many areas such as computer vision, image processing and graphics. For instance, it has been successfully applied in object recognition \cite{Ren2014Region}, hyperspectral image classification \cite{wang2016salient}, object tracking \cite{Borji2012Adaptive}, image compression \cite{fang2012saliency}, and  image resizing \cite{battiato2014saliency, he2015supercnn}. As a pre-processing step, salient object detection methods should process in real time which is an essential application requirement.

It is very challenging to automatically, efficiently and accurately estimate salient objects without any actual scene understanding, especially in complex scenes. Inspired by the cognitive theories and physiological models of visual attention, researchers in computer science propose lots of saliency detection models based on low-level  visual information.
\begin{figure}[!tbp]
	\centering
	\includegraphics[width=3.4in]{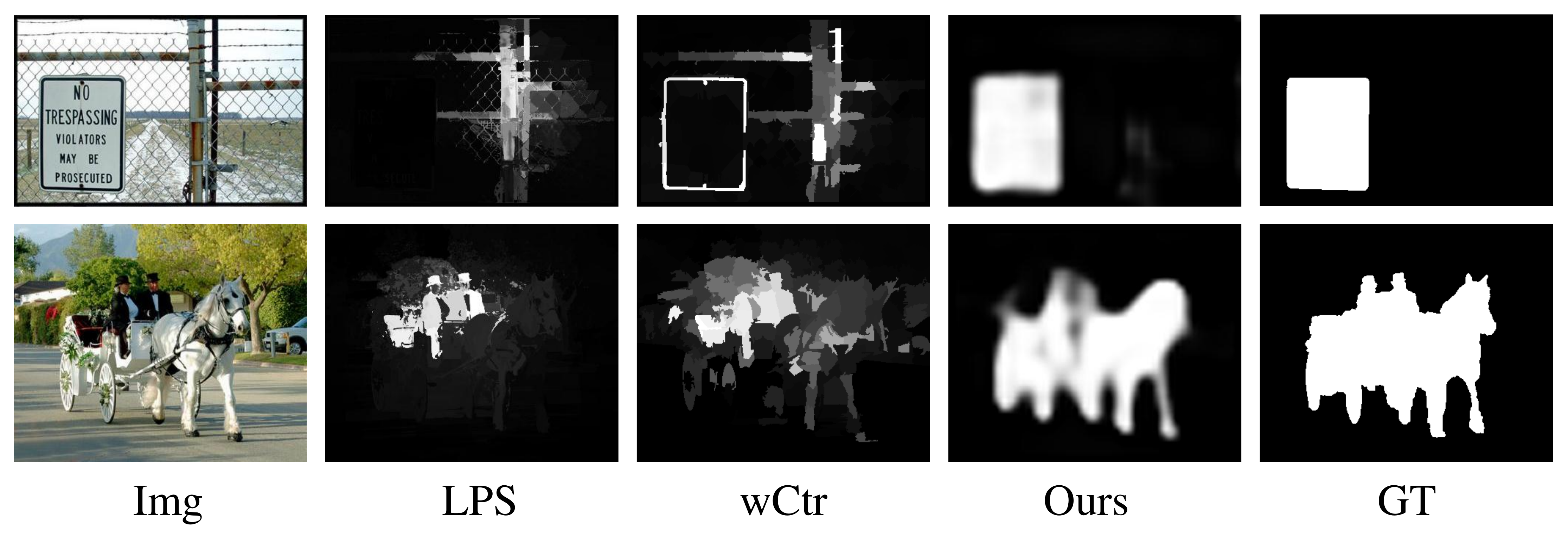}
	\caption{Saliency detection results based on low-level priors and high-level semantic features. From left to right: original image (Img), saliency detection based on low-level priors (LPS \cite{li_tip15_LPS}, wCtr \cite{2014CVPR_RBD}), our method based on high-level semantic features (Ours) and ground truth (GT).}\label{examples}
\end{figure}
Traditionally, saliency detection methods leverage various low-level saliency priors such as contrast prior, compactness prior, objectness prior, background prior and center prior to estimate saliency. These low-level saliency priors get impressive performances in the state-of-the-art saliency detection methods. However, in the case of complex scenes or when salient objects are semantically salient in an image, low-level priors can hardly make salient objects pop out from an image, and these methods may be fragile and fail. For example, LPS \cite{li_tip15_LPS} fuses boundary prior and objectness prior into inner and inter label propagation scheme for saliency estimation, and wCtr \cite{2014CVPR_RBD} uses boundary connectivity to measure the backgroundness for saliency inference. While these low-level saliency priors can not make the signs pop out from the images, and these methods fail in these complex scenes, as shown in \figurename{\ref{examples}}.

Due to the shortcomings of low-level saliency priors, many methods try to incorporate high-level visually semantic concepts into saliency detection procedure, which is also important for human to estimate the saliency in complex scenes. Because deep convolutional neural networks (CNN) are good at extracting high-level semantic information, which has achieved better performances than previous methods based on handcrafted features, it is natural to consider convolutional neural networks for saliency detection. Recently, many saliency detection models based on deep learning have been proposed \cite{li2015visual_MDF,MCDL2015CVPR,wang2015deep_LEGS,ELD2016CVPR,DCL2016CVPR}. These deep learning based saliency detection models either use convolutional networks to extract hierarchical features for saliency estimation or use hierarchical networks to infer saliency score. Although these methods achieve state-of-the-art performances, they always include complex pre-processing or post-processing procedures for better performances, and can not implement an end-to-end learning for saliency detection. Complicated and time-consuming procedures make them not easy to be practically applied as a pre-processing step in computer vision tasks.

In order to get a computational-efficiency salient object detection method and make the procedure fast and compact, we propose a novel end-to-end salient object detection method which implements saliency score regression based on a single fully convolutional network (FCN) without any pre-processings and post-processings. As for the network architecture, it replaces top fully connected layers of the famous VGG-16 Net with fully convolutional layers, which can achieve a good balance between the precision and speed on semantic perception. These fully convolutional layers together play a role of nonlinear regression from feature maps to pixel-level saliency scores. Considering the characteristic of the salient object detection task, we utilize a single full-resolution input manner which can make images keep natural visual information. For the sake of not increasing too much computation, we employ a bi-cubic interpolation method to exactly restore the size of saliency maps, and encapsulate the method in an independent layer. In addition, we design a suitable loss function to guide the network to converge to a good solution for the task. After training, the network can directly predict a dense full-resolution saliency map when given an image. It works like a compact pipeline which effectively simplifies the processing procedure. We compare the proposed method with 10 state-of-the-art methods on 6 public datasets, and the experimental results show that the proposed method can achieve comparable or better precision performance while get a significant improvement in processing speed (35 FPS). It ensures that our method can run in a real time manner and can be practically used as a pro-processing step before other visual tasks.

In total, this paper carries four major contributions summarized below.

\begin{enumerate}[1)]
	\item
	For the salient object detection task, we propose a end-to-end deep neural network which can directly predict dense full-resolution saliency maps from original images without any pre-processings and post-processings (image-to-image prediction).  It works like a compact pipeline which extremely simplifies the detection procedure comparing with contemporary deep learning based methods. (Section \ref{section3}) 
	\item
	We further promote the representation ability of CNN for the salient object detection task by adopting a single full-resolution input manner and a specially designed loss function. (Section \ref{section3})
	\item
	We evaluate our method quantitatively and qualitatively with comprehensive experiments, and experimental results show that our method can achieve comparable or better precision performance than the state-of-the-art methods while get a significant improvement in the detection speed (processing in real time). (Section \ref{comparisionExp})
	\item
	We verify the effectiveness of the single full-resolution input manner and the specially designed loss function for the salient object detection task through controlled experiments. (Section \ref{AnalysisPart})		
\end{enumerate}
 	
\section{Related Work}
%
%
%
%
%
%
Recently, deep convolutional neural network has showed its powerfulness in feature representation, which achieves substantially better performance than previous state-of-the-art methods in many computer vision tasks, including salient object detection. In this section, we will review recent salient object detection methods based on deep learning.

As deep convolutional neural network can build hierarchical architecture to extract high-level features of an image, which is important for saliency detection, many salient object detection methods adopt CNN to extract high level features. For instance,  Zou \cite{zou2015harf} incorporates multi-layered deep learning features from multi-level regions as elementary features into a hierarchy-associated feature construction framework for salient object detection. Zhao \cite{MCDL2015CVPR} proposes a multi-context deep learning framework for salient object detection, which takes global context and local context into account based on superpixel segmentation. Li \cite{li2015visual_MDF} incorporates multiscale CNN features extracted from nested windows with deep convolutional neural networks for saliency estimation, and integrates the CNN-based saliency model with spatial coherence model and multi-level image segmentation \cite{jiang2013salient_MSRAB}. Wang \cite{wang2015deep_LEGS} uses a deep neural network to learn local patch features to determine saliency score, and joints global contrast and geometric information to describe object candidate regions which are generated by geodesic object proposal \cite{krahenbuhl2014geodesic}. It uses the regions in sliding windows to evaluate their saliency, which may result in the salient object and background in the same sliding window having the same saliency. Lee \cite{ELD2016CVPR} constructs a low-level distance map based on hand-crafted low-level features for superpixel regions produced by SLIC, and encodes it with a convolutional neural network. Then this encoded low-level distance map and high-level features extracted by the VGGNet \cite{VGG} are concatenated to evaluate the saliency of a query region. Tang \cite{tang2016saliency} combines pixel-level saliency estimation and  region-level saliency estimation for saliency detection with CNNs. Kim \cite{kim2016shape} combines a shape prediction driven by a convolutional neural network with mid- and low-regions preserving image information for salient object detection. In this method, it adopts the selective search \cite{uijlings2013selective} to extract category independent region proposals. All these methods are based on superpixel segmentation or object proposal extraction, which consume much time to perform the region segmentation, and some generated regions are under-segmented or over-segmented. Meanwhile, these region generation procedures make the methods complex and not to be end-to-end saliency detection models.

Fully convolutional network \cite{chen2014semantic, long2015fully, xie2015holistically, noh2015learning} trained end-to-end, pixels-to-pixels, exceed the state-art-of-the-art methods in semantic segmentation, which motivates recent research efforts of using fully convolutional neural networks for salient object detection \cite{liu2016dhsnet, li2016deepsaliency, DCL2016CVPR, Kruthiventi_2016_CVPR, Kuen_2016_CVPR, wang2016saliency}. Liu \cite{liu2016dhsnet} proposes an end-to-end deep hierarchical saliency network based on convolutional neural networks. It firstly makes a coarse global saliency prediction by learning various global structured saliency cues and their optimal combination, and then refines saliency maps with a hierarchical recurrent convolutional neural network step by step via integrating local context information.  Li \cite{li2016deepsaliency} designs a multi-task deep saliency model based on a fully convolutional neural network with global input and global output, which takes a data-driven strategy for encoding underlying saliency prior information and sets up a multi-task learning scheme for exploring the  intrinsic correlations between saliency detection and semantic image segmentation. Li \cite{DCL2016CVPR} proposes an end-to-end deep contrast network for salient object detection, which consists of two complementary components, a pixel-level fully convolutional stream and a segment-wise spatial spooling stream. Kruthiventi \cite{Kruthiventi_2016_CVPR} proposes an unified framework based on deep convolutional architecture for predicting eye fixations and segmenting salient objects.  It designs the initial network layers, shared between both tasks, such that they can capture the object-level semantics and the global contextual aspects of saliency, while the deeper layers of the network address task specific aspects. Kuen \cite{Kuen_2016_CVPR} designs a recurrent attentional convolutional-deconvolutional network to refine saliency maps generated by convolutional-deconvolutional network \cite{noh2015learning}, which can perform salient object detection in an end-to-end fashion. Wang \cite{wang2016saliency} develops a recurrent fully convolutional network for saliency detection, which can incorporate saliency prior knowledge for more accurate inference and automatically learn to refine a saliency map by correcting its previous errors. These methods always include a preparation stage which warps irregular images to an uniform size in the model training and testing phases. The warp operation may corrupt semantic contexts of images and change structural characteristic of objects in images, that may degrade the model performance.

Comparing with the aforementioned methods, our method adopts an extremely simplified end-to-end deep neural network without any pre-processings and post-processings. It works like a compact pipeline which extremely simplifies the detection procedure. It can achieve comparable or better precision performance than the state-of-the-art methods while get a significant improvement in the detection speed (processing in real time).

\section{Saliency Score Regression Network}\label{section3}

Our aim is to design a fast and compact salient object detection method to meet the essential application requirement of this task. Driven by this target, we propose a saliency score regression network which is based on a single fully convolutional network. Beyond the network, there is no additional pre-processing and post-processing procedures. We design the network mainly from two aspects: network architecture and loss function. In this section, we describe the proposed method from these two aspects in detail.

\subsection{Network Architecture Design}\label{structure}

\begin{figure*}[!t]
	\centering
	\includegraphics[scale=0.21]{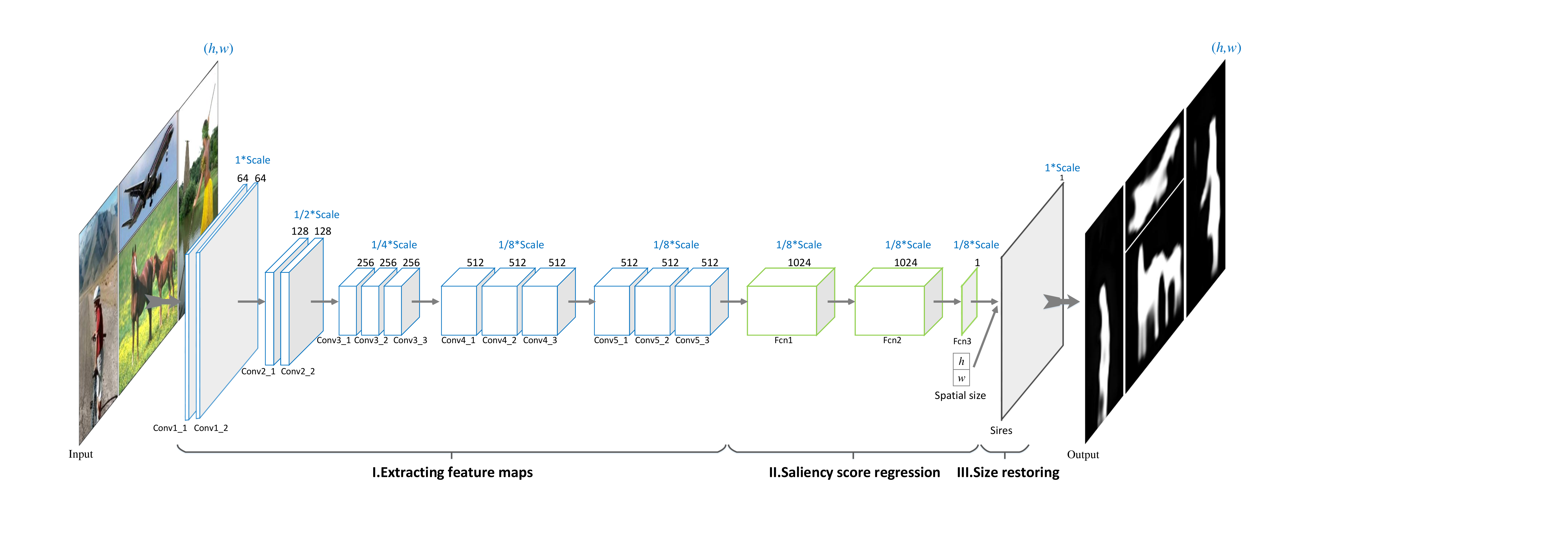}
	\caption{The framework of our proposed method. One single arbitrary-size image is fed into the network, and then the network directly outputs a full-resolutiuon saliency map. There is no additional pre-processing and post-processing beyond the network.}
	\label{framework}
\end{figure*}

The mapping relationship from an original image to a saliency map is complicated, and the network must have enough hidden units to model it accurately. Meanwhile, we have to take in account of the computation complexity achieving real-time processing and end-to-end processing manner. In recent years, deeper convolutional network architectures achieve better performances while cost more computation time \cite{AlexNet2012,VGG,Googlenet,ResNet,Szegedy2016Inception}. After referring to these famous networks, we consider that the famous VGG16-Net\cite{VGG} is a better balance between the speed and precision, and we regard it as a base of our method. Our saliency score regression network is constructed by modifying the VGG16-Net, and its brief architecture is illustrated in Fig.~\ref{framework}. It replaces 3 upper connected layers of the VGG16-Net for image classification with 3 fully convolutional layers ($1\times1$  convolution filters) and adds a size restoration layer. In total, it has 17 layers. For the sake of clear explanation, immediate feature maps at each layer are denoted as $X_i,i=1,...,17$ ($X_{17}$ is the final output). In addition, we adopt an input manner based on single full-resolution images. The whole network can be divided into three functional parts: (\uppercase\expandafter{\romannumeral1}) extracting feature maps, (\uppercase\expandafter{\romannumeral2}) saliency score regression, (\uppercase\expandafter{\romannumeral3}) size restoring. The details are explained below.

\subsubsection{Extracting Feature Maps}\label{Architecture1} 

This functional part is consist of 5 convolutional stages and is responsible for extracting dense pixel-level features. The input is original image $X$ and the outputs are feature maps $X_{13}$. Below we describe the input manner and these convolutional stages orderly.     

At the entrance of the proposed network, we adopt a novel input manner which is different from existing salient object detection method based on FCN. Firstly, the input is only a normal color image, and no superpixel-level or region-level segmentation map is fed into our network. In \cite{ELD2016CVPR,DCL2016CVPR,tang2016saliency}, color images and superpixel-level or region-level segmentation results are all fed into networks. Though these low-level pre-processings can, to some extent, guarantee coherence and connectivity, the model complexity and computation time always decrease the practicability. Considering the essential application requirement of salient object detection task, we abandon the advantage of these low-level pre-processings. This way can effectively simplify the whole processing procedure and reduce the computation time. Secondly, our method inputs single full-resolution (full-size) images to the network. Saliency estimation in an image is based on global context (including a full image) which can offer complete information to model the saliency between all the objects. Most existing saliency detection methods based on FCNs\cite{ELD2016CVPR,MCDL2015CVPR,li2015visual_MDF} input a batch of uniform-size full (not full-resolution) images to their network at the training phase. At the preparation stage, they have to warp irregular training images to a uniform size to make their network work. However, the warp operation may cause several problems: a) corrupting semantic contexts of images; b) changing the characteristic of objects in images involved with dimension (such as length-width ratio); c) reducing the resolution of images which especially increase the difficulty of detecting small objects. These problems may slightly puzzle neural networks when perceiving images and lower corresponding performances. The full-resolution input manner can keep away from these problems, and ensure that input images are perceived with natural contexts. However, the network processes only one image during both training and test phase to achieve the full-resolution input manner. This way makes the network more sensitive to the distribution of training samples and makes the network harder to converge. Luckily, we can design a smooth and robust loss function to overcome the drawback and guide the network to converge to a good solution. The details of our loss function will be explained in the next subsection.    

After entering the network, an input image is processed by a stack of 5 convolutional stages which respectively contain 2, 2 ,3, 3 and 3 convolutional layers (denoted by $Convi\_j, i \in \{ 1,2,3,4,5\} ,j \in \{ 1,2,3\}$). These convolutional layers all adopt a very small receptive field: $3\times3$ kernel size which is the smallest size to capture the notion of neighbor. The numbers of features maps produced by these 13 convolutional layers are adopted as \cite{VGG} suggested.  The detailed numbers are illustrated in Fig.~\ref{framework}. To increase the nonlinear representation ability, each of these convolutional layers is followed by a rectified linear unit layer denoted by $Relu$ (for simplicity, it is not shown in Fig.~\ref{framework}) which can make the network converge more faster than traditional nonlinear activation function \cite{AlexNet2012}. Meanwhile, each convolutional stage is followed by a max-pooling layer with $3\times3$ kernel. As for the strides of these five max-pooling layers, the first three are set to 2 while the others are set to 1. Then throughout this function part, the size of outputs $X_{13}$ becomes ${\raise0.7ex\hbox{$1$} \!\mathord{\left/{\vphantom {1 8}}\right.\kern-\nulldelimiterspace}\!\lower0.7ex\hbox{$8$}}$ of input images $X$.

\subsubsection{Saliency Score Regression} This functional part contains a cascade of 3 fully convolutional layers  (denoted by $Fcni, i \in \{ 1,2,3\}$) and can be viewed as a nonlinear regression from feature maps extracted by the ahead functional part to pixel-level saliency scores. Each fully convolutional layer is followed by a $Relu$ layer and a $Dropout$ layer. As it is fairly expensive to label salient objects in images, existing salient object detection datasets are all in a small amount which can easily lead the training to be overfitting. The $Dropout$ layer has been proven to produce overfitting effectively by omitting the hidden units from the network with probability $\alpha$ \cite{HintonDropout} (0.5 is common and adopted in this paper).

\subsubsection{Size Restoring}  

This function part contains only one layer $Sires$ and is responsible for restoring the size of saliency maps. In fact, the network has produced a basic saliency map $X_{16}$ after the regression operated by these 3 fully convolutional layers. Although $X_{16}$ has exactly detected salient objects in the input image $X$, the size of $X_{16}$ is around (not exactly, as the size of input image is arbitrary ) ${\raise0.7ex\hbox{$1$} \!\mathord{\left/{\vphantom {1 8}}\right.\kern-\nulldelimiterspace}\!\lower0.7ex\hbox{$8$}}$ of $X_0$. For the sake of not increasing too much computation, we abandon the benefits taken by deconvolutional layers and employ a bi-cubic interpolation method over $4\times4$ pixel neighborhood to exactly restore the size of saliency maps based on the wide $w$ and height $h$ of the input image $X$. $w$ and $h$ are recorded when $X$ is fed into the network, and then are transmitted to the size restoring layer. We encapsulate the method in an independent layer and name it as size restoration layer (denoted by $Sires$). After the size restoring, the network can produce the final result $X_{17}$, a full-resolution saliency map. The results produced by this method are smoother and have fewer interpolation artifacts than the ones produced by bilinear interpolation and nearest-neighbor interpolation method.

Of course, the simple size restoring method can not preserve fine edge details of salient objects. We can overcome the drawback by some approaches such as CRF. In fact, the CRF is widely used in recent works and effectively improves the spatial coherence of detection results \cite{DCL2016CVPR}. We also tried to connect a CRF layer to our network and found that it can further improve the performance by more than $1.0\%$ while increase extra $50+$ ms in runtime for each picture. To guarantee the characteristics of processing in real time, we abandon the potential approach making our precision performance better. 

In above paragraphs, we describe our network architecture detailedly. It is an extremely simplified end-to-end deep neural network. When given an image, the trained network can directly predict a dense full-resolution saliency map without any pre-processings and post-processings. It works like a compact pipeline which effectively simplifies the processing procedure and makes the method easy to be practically applied. 

In addition, the parameters learned in training phase are from convolutional layers and fully convolutional layers. We can infer that the total number of learned parameters in the model is about 19 million. The number is smaller than the ones of most state-of-the-art methods. We show the advantage in the Section \ref{AnalysisC}.

\subsection{Loss Function Design} \label{lossdesign}

In above subsection,  we illustrate the architecture of the proposed method. To make the network work as we expect, we need to design a suitable loss function to guide the work to converge to a good solution. The design bases are explained below. 

Firstly, as for the salient object detection task, samples can be used in training are scarce for deep learning based methods which always require large amount of training samples to learn latent rules. However, the visual task has to face more objects categories and more factors affecting the final results compared with a similar task semantic segmentation (such as color, position and scene). It needs more samples to represent the essential sample space. However, existing publicly available datasets are all at the magnitude of thousands. Comparing with famous large-scale datasets \cite{ImageNetDataset,COCOMicrosoft} for other visual tasks (such as image classification, object detection and scene parsing), they are all lightweights. What's more, during the scarce samples, the numbers of salient and non-salient pixels are heavily imbalanced. Comparing with variety of non-salient objects, salient objects are often in a smaller amount. To make full use of valuable salient objects, the total loss function should weight the loss caused by the pixels of salient objects in groud truth. Considering these elements, we design the total loss function represented by 
\begin{equation}
\label{totalloss}
L(X,Y,\theta ,\beta ) = \frac{1}{N}\left[ {{L_g}(X,Y,\theta ) + \beta  * {L_s}(X,Y,\theta )} \right],
\end{equation}
where $X$ is an input image, $Y$ is corresponding target saliency map, $\theta$ is the network parameter, $\beta$ is a super-parameter and $N$ is the pixel number of a whole image. The total loss function consists of two items, basic global loss ${L_{g}}(X,Y,\theta )$ and extra salient-region loss ${L_{s}}(X,Y,\theta )$. The super-parameter $\beta$ is used to balance the two items. The detailed expression of the two items are explained below. Additionally, it should be noticed that the weighted sum of the two items is averaged by the number of pixels. The average operation can make the loss value not be affected by the image size.

In above subsection, we explains the reason why we adopt the input manner based on single full-resolution images, and corresponding negative effect that makes the network more sensitive to the distribution of training samples and makes the network harder to converge. In order to overcome the drawback, we design a smooth and robust loss function during the forward and backward propagation. The two sub-items in Eq.(\ref{totalloss}) are embodied as
\begin{equation}
{L_g}(X,Y,\theta ) = \sum\limits_{i = 1}^N {\Psi \left[ {F\left( {{x_i},\theta } \right) - {y_i}} \right]}, 
\end{equation}
\begin{equation}
{L_s}(X,Y,\theta ) = \frac{{{N^ - }}}{N}\sum\limits_{j = 1}^{{N^ + }} {\Psi \left[ {F\left( {{x_j},\theta } \right) - {y_j}} \right]},
\end{equation}
where $N^+$ is the pixel number of salient objects,  $N^-$ is the pixel number of non-salient objects, $x_i$ represents a pixel in an input image $X$ ($X=\{ x_i \ | \ i=1,2,...,N \}$), $y_i$ represents a pixel in a target saliency map $Y$ ($Y=\{ y_i \ | \ i=1,2,...,N \}$, $y_i \in \{ 0,1\} $), function $F(\cdot)$ represents an abstraction of the network's processing, $\Psi(\cdot)$ is a robust loss function (Smooth-L1) reported in \cite{FastRCNN} which is explained in detail below, and other existing symbols have the same meanings as in Eq.(\ref{totalloss}). $F\left( {{x_i},\theta } \right)$ represents the regression value of saliency score for each pixel. ${L_{g}}(X,Y,\theta )$ represents a basic loss from all the pixels in an image. It forces the regression results to approach the ground truth from a global scope. ${L_{s}}(X,Y,\theta )$ represents an extra loss from the pixels belonging to salient objects. It makes the loss function pay more attention to the region of salient objects than background.

The Smooth-L1 function $\Psi(\cdot)$ is defined as
\begin{equation}
\Psi \left( z \right) = \left\{ {\begin{array}{*{20}{c}}
	{0.5{z^2},{\rm{ if }}\left| z \right| \le 1}\\
	{\left|z \right| - 0.5,{\rm{otherwise.}}}
	\end{array}} \right.
\end{equation}
It can be easily inferred that the Smooth-L1 is smooth and derivable in the real field. Its derivation is
\begin{equation}
\Psi^{'} \left( z \right) = \left\{ {\begin{array}{*{20}{c}}
	{z,{\rm{if }}\left| z \right| \le 1}\\
	{{\rm{sign}}\left( z \right),{\rm{otherwise.}}}
	\end{array}} \right.
\end{equation}
Compared with common L1 loss function, the derivation of Smooth-L1 is continuous at the zero point. It can make the network converge more stably near the zero point. In the interval $[ - 1, + 1]$, $\Psi^{'} \left( z \right)$ is directly proportional to the error $z$. It can make the network converge with a self-adaptive step and stabilize at a locally optimal solution. Outside of the interval,  $\Psi^{'} \left( z \right)$ is truncated to $\pm 1$. It can make the converging procedure more robust and less affected by singular samples. During the training phase, we utilize standard stochastic gradient descent (SGD) algorithm to minimize Eq.(\ref{totalloss}).

Above is a theoretical analysis of our loss function. We will show its advantage through a controlled experiment in Section \ref{AnalysisB}.

\section{Experiments and Analyses}

In this section, we verify the effectiveness of our method through comprehensive experiments. We evaluate our method quantitatively and qualitatively on 6 public datasets, and we also compare our method with 10 recent state-of-the-art methods. We implement the comparison under common evaluation metrics which can dig the characteristics of our method from all angles. In addition, we analyze our method from 3 aspects to lighten potential cues supporting our advantages.

\subsection{Datasets}
We evaluate our method on 6 public datasets: DUT-OMRON \cite{yangCVPR2013_GMR}, PASCAL-S \cite{secrets2014li_PASCALS}, ECSSD \cite{CVPR2013_HS}, HKU-IS \cite{li2015visual_MDF}, SED2 \cite{4270042_SED2} and MSRA-B\cite{MSRA-B}. All the datasets have pixel-wise ground truth annotations. The DUT-OMRON has 5,168 high quality images which are manually selected from more than 140,000 natural images. Each image has one or more salient objects and a relatively complex background. The PASCAL-S contains 850 natural images which are built on the validation set of the PASCAL VOC 2010 segmentation challenge. It is a challenging saliency dataset as many images have highly cluttered backgrounds and multiple complex foreground objects. We threshold original salient object annotations at 0.5 to obtain binary masks as suggested in \cite{secrets2014li_PASCALS}. The ECSSD contains 1,000 semantically meaningful and structurally complex images, which have multiple objects of different sizes. HKU-IS is another large dataset and contains 4,447 images, most of which have either low contrast or multiple salient objects. SED2 contains 100 images and each of them has two salient objects. MSRA-B has 5,000 images and is widely used for the salient object detection task. Most of the images contain only one salient object. 

\subsection{Evaluation Metrics}

We evaluate the performance of our method with various metrics \cite{SalObjBenchmark}: precision-recall (PR), receiver operating characteristics (ROC), area under ROC curve (AUC), F-measure of average precision recall curve, mean absolute error (MAE) score, and computation time (runtime).

\textbf{Precision-recall (PR).} The PR curve is obtained by binarizing the saliency map $S$ to binary masks $M$ with a set of fixed thresholds in $[0,1,...,255]$ and comparing them with ground truth $G$. The precision refers to the fraction of salient pixels which are assigned correctly in the detected saliency maps. While the recall refers to the fraction of correct salient pixels in the ground truth: 
\begin{equation}\label{Precision}
Precision = \frac{|M \cap G|}{|M|}, Recall = \frac{|M \cap G|}{|G|}.
\end{equation}
All the precision and recall scores are combined to plot the PR curve. 

\textbf{Receiver operating characteristics (ROC).} The ROC curve is generated based on true positive rates (TPR) and false positive rates (FPR) when binarizing saliency maps with a set of fixed thresholds:
\begin{equation}\label{ROC}
TPR = \frac{|M \cap G|}{|G|}, FPR = \frac{|M \cap \bar{G}|}{|\bar{G}|},
\end{equation}
where $\bar{G}$ denotes the oppositive of the ground truth $G$. The ROC curve  plots the $TPR$ versus $FPR$ by varying the threshold.

\textbf{Area under ROC curve (AUC). } The AUC score is computed as the area under the ROC curve. A perfect AUC performance gets a score of 1, while the AUC performance of random guessing gets a score of 0.5.

\textbf{F-measure.} It evaluates a binarized map with respect to ground truth based on precision and recall. It is defined as a weighted harmonic mean of precision and recall with a non-negative weight $\beta$:
\begin{equation}\label{F-measure}
F_\beta = \frac{(1+ \beta^2)Precision \times Recall}{\beta^2 Precision + Recall},
\end{equation}
where we set $\beta^2=0.3$ as in \cite{SalObjBenchmark} to give more importance to precision. In order to get a good summary of the detection performance, we use the maximal $F_\beta$ to score the PR curve produced by fixed thresholding.

\textbf{Mean absolute error (MAE).} The MAE score is calculated as the mean of pixel-wise absolute errors between the saliency map  $S$ and the ground truth $G$:
\begin{equation}\label{MAE}
MAE = \frac{1}{W_I \times H_I} \sum_{x=1}^{W_I}\sum_{y=1}^{H_I} | S(x,y) - G(x,y)|,
\end{equation}
where $W_I$ and $H_I$ are the width and height of the saliency map $S$. $S(x,y)$ and $G(x,y)$ are the continuous saliency score and the binary ground truth at pixel $(x,y)$, which are normalized in the range $[0,1]$. Smaller MAE score means better performance.

\textbf{Runtime.} We not only evaluate all the methods in terms of precision but also evaluate their runtime. For fair comparisons, all the codes run at the same personal computer (PC) which owns a single NVIDIA GeForce GTX 1080P GPU with 8GB memory and Intel Core i7-4770 @3.4 GHz. In principle, the runtime counts the time cost by the main detection procedure (not including the I/O time) for each image, which can indeed reflect computation cost. As the operation of reading images is often coupled with the main detection procedure, it is not easy to separate reading images out when counting runtime. In this paper, the runtime starts when reading an image from a disk and ends when obtaining a full-resolution detection result (not including the time used for writing a result map to a disk). In addition, all the methods are not allowed to process multiple images in parallel and are fed with full-resolution images. We present mean runtimes on corresponding datasets below.

\subsection{Implementation}

\begin{table*}[!t]
	\caption{Code types of all the comparison methods.}
	\label{CodeType}
	\centering
	\begin{tabular}{ c | c c c c c c c c c c c }
		\toprule
		Method  &GMR    &HS    &wCtr    &MB     &LPS     &MDF      &MCDL  &LEGS  &ELD   &DCL   &Ours  \\ 
		\hline
		Code Type	&Matlab  &Exe   &Matlab  &Exe    &Matlab  &Caffe    &Caffe  &Caffe &Caffe &Caffe &Caffe \\ 
		\bottomrule
	\end{tabular}
\end{table*}

\textbf{Training data.} The training data significantly affects the final behavior of a deep neural network \cite{DCL2016CVPR}. As in \cite{ELD2016CVPR,wang2016saliency}, we train our model on the MSRA10K\cite{CVPR2007MSRA10K} dataset. The dataset has 10,000 images and most of them contain single object. It can offer more samples and more simplex sample distributions than other large datasets. A little different from them, we employ some samples to validate the model at a fixed period. That is because the loss, under the single-image input manner, still varies even though the model has converged. To enlarge the amount of training samples as possible, we do not reserve testing samples. In detail, we randomly divide the MSRA10K dataset into two subsets, 8000 samples for training and 2000 samples for validation.

Additionally, we utilize the trick data augmentation which has been proven to improve performance effectively in many learning-based vision tasks \cite{BestPractices,Yaeger2010Effective,AlexNet2012}. There are many approaches to augment training data such as rotation, skewing, warping and flipping. However, artificially synthesizing samples must be reasonable for a certain problem. In the case of salient object detection task, sample distribution has some invariance with respect to horizontal-flipping from an intuitive viewpoint. Then we flip all the training images horizontally, which results in an augmented training set which is twice larger than the original one.


\textbf{Parameters.} Our method is build on top of the publicly available implementation of FCN \cite{CP2016Deeplab} which uses the Caffe library. We complete all the computation of our method under the framework.  In the training phase, we adopt "step" learning rate policy and set base learning rate to 0.01, step size to 3000, gamma (multiplier) to 0.1. Other hyper-parameters used in the training include: momentum (0.9), weight decay (0.0005), and maximum iteration (12000). The entire training procedure takes 9 hours on the PC mentioned above.

\textbf{Fine-tuning.} It is difficult to train a deep network from scratch when training samples are not enough. As existing saliency datasets are all in a small amount, training our network on them directly from an initial state is inadvisable. As verified in many works, fine-tuning can provide a good initialization for training deep models and make the training more effective. The training of our model is fine-tuned from a pre-trained model, DeepLab-LargeFOV \cite{CP2016Deeplab} for semantic segmentation task, which is a variant of DeepLab with faster training time. In addition, to test the adaptability and make a fair evaluation, our model is directly evaluated on the 6 datasets without fine-tuning.

\subsection{Comparison with State-of-the-art Methods} \label{comparisionExp}

We compare the proposed method with 10 recent state-of-the-art methods on aforementioned datasets, which include GMR \cite{yangCVPR2013_GMR}, HS \cite{CVPR2013_HS}, wCtr \cite{2014CVPR_RBD}, MB \cite{zhang2015_MB}, LPS \cite{li_tip15_LPS}, MDF \cite{li2015visual_MDF}, MCDL \cite{MCDL2015CVPR}, LEGS \cite{wang2015deep_LEGS}, ELD \cite{ELD2016CVPR} and DCL \cite{DCL2016CVPR}. The first 5 methods are classical unsupervised methods, while the last 5 methods are deep learning based methods. For fair comparison, we practically run all the codes with the parameters offered by authors. Their code types are summarized in Table \ref{CodeType}. Note that, the five deep learning based methods and our method are all implemented with the Caffe library. In below experiments, they are evaluated without acceleration by cuDNN. 

We present the comparison results from both qualitative and quantitative aspects for comprehensively revealing the characteristics of our method.

\subsubsection {Qualitative Performance Comparison}

\begin{figure*}[!t]
	\centering
	\includegraphics[scale=0.245]{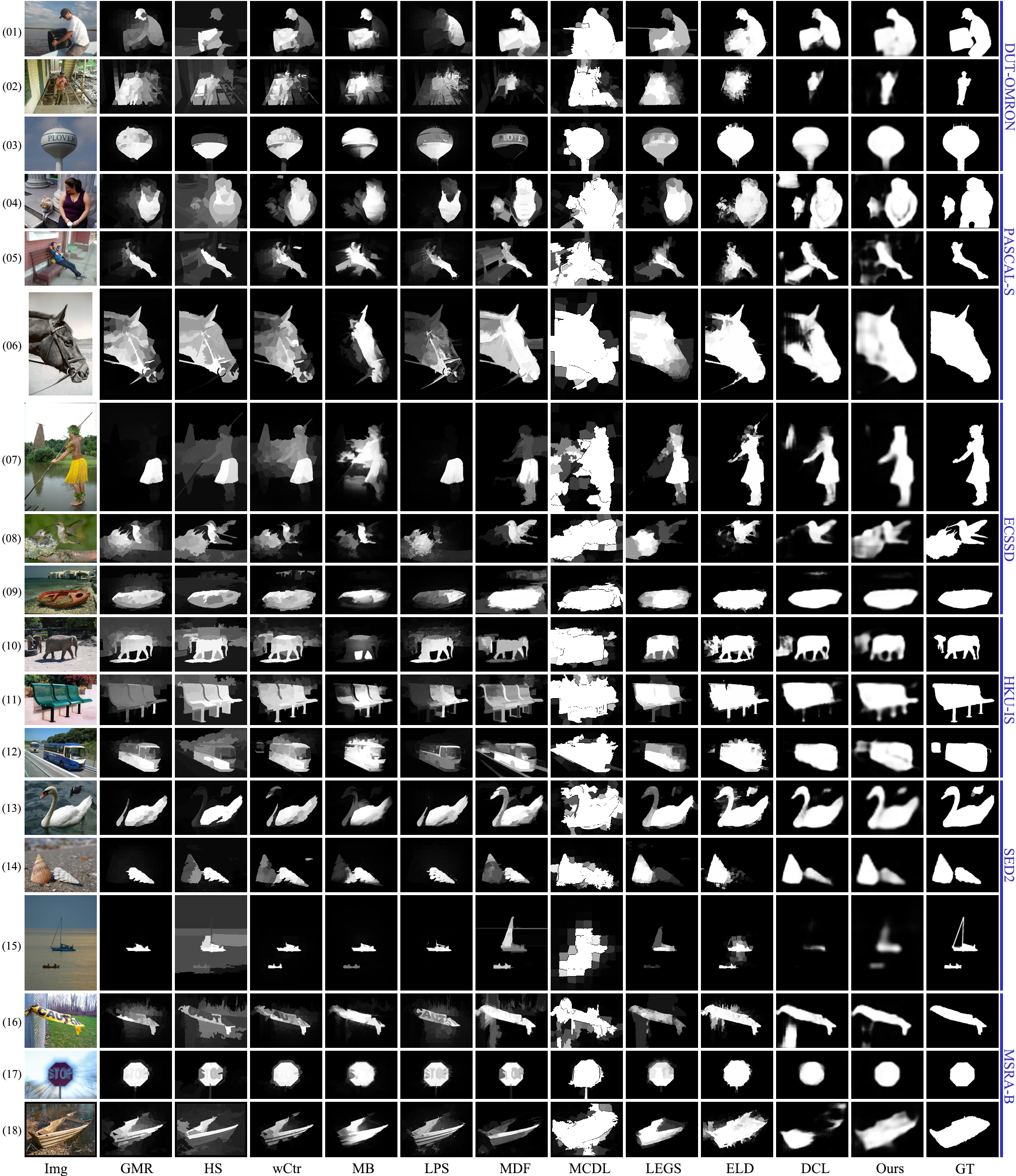}
	\caption{Qualitative comparison of different methods with oringinal image (Img) and ground truth (GT) on the DUT-OMRON, PASCAL-S, ECSSD, HKU-IS, SED2 and MSRA-B datasets. 3 challenging samples are taken from each dataset and are arranged successively.}
	\label{saliencyMaps}
\end{figure*}

For an intuitive illustration, in Fig.~\ref{saliencyMaps}, we provide some detection results of the proposed method and aforementioned 10 methods on the 6 datasets. We only select 3 challenging samples from each dataset for the space limit. It can be seen from the figure that: a) our method is good at detecting semantically salient objects even though the targets are small (i.e. Row 12 and 15); b) our method can detect the most salient objects from highly cluttered backgrounds (i.e. Row 2 and 18); c) our method can give more clean (i.e. Row 7 and 16), intact (i.e. Row 8 and 14) and smooth  (i.e. Row 9 and 13) saliency maps. In total, our method can create more visually favorable saliency detection results. However, we admit that our method is unable to capture tiny details for adopting a simple size restoring approach to keep less runtime.

\subsubsection {Quantitative Performance Comparison}

\begin{figure*}[!t]
	\centering
	\subfigure{\includegraphics[width=5.9cm]{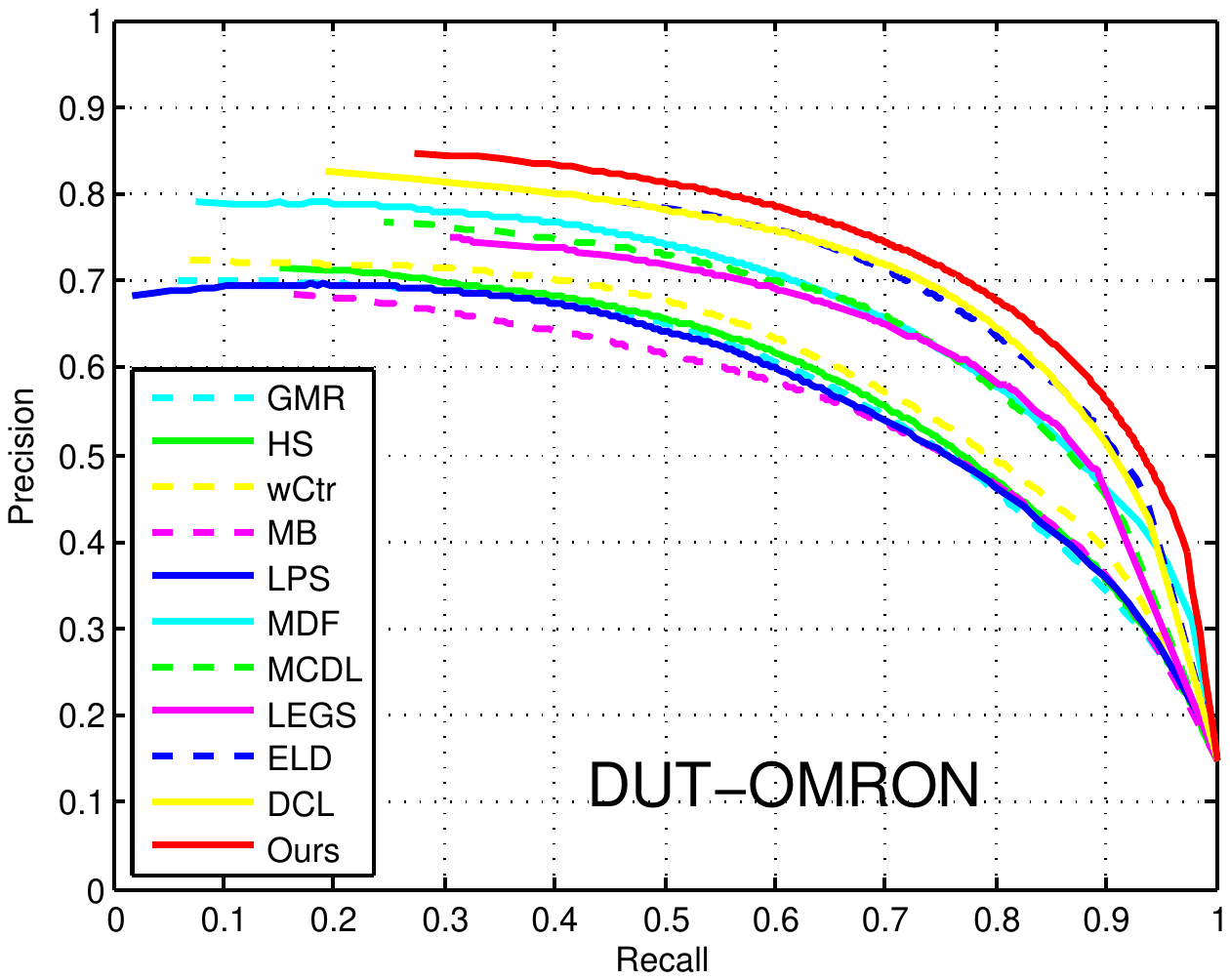}}  
	\subfigure{\includegraphics[width=5.9cm]{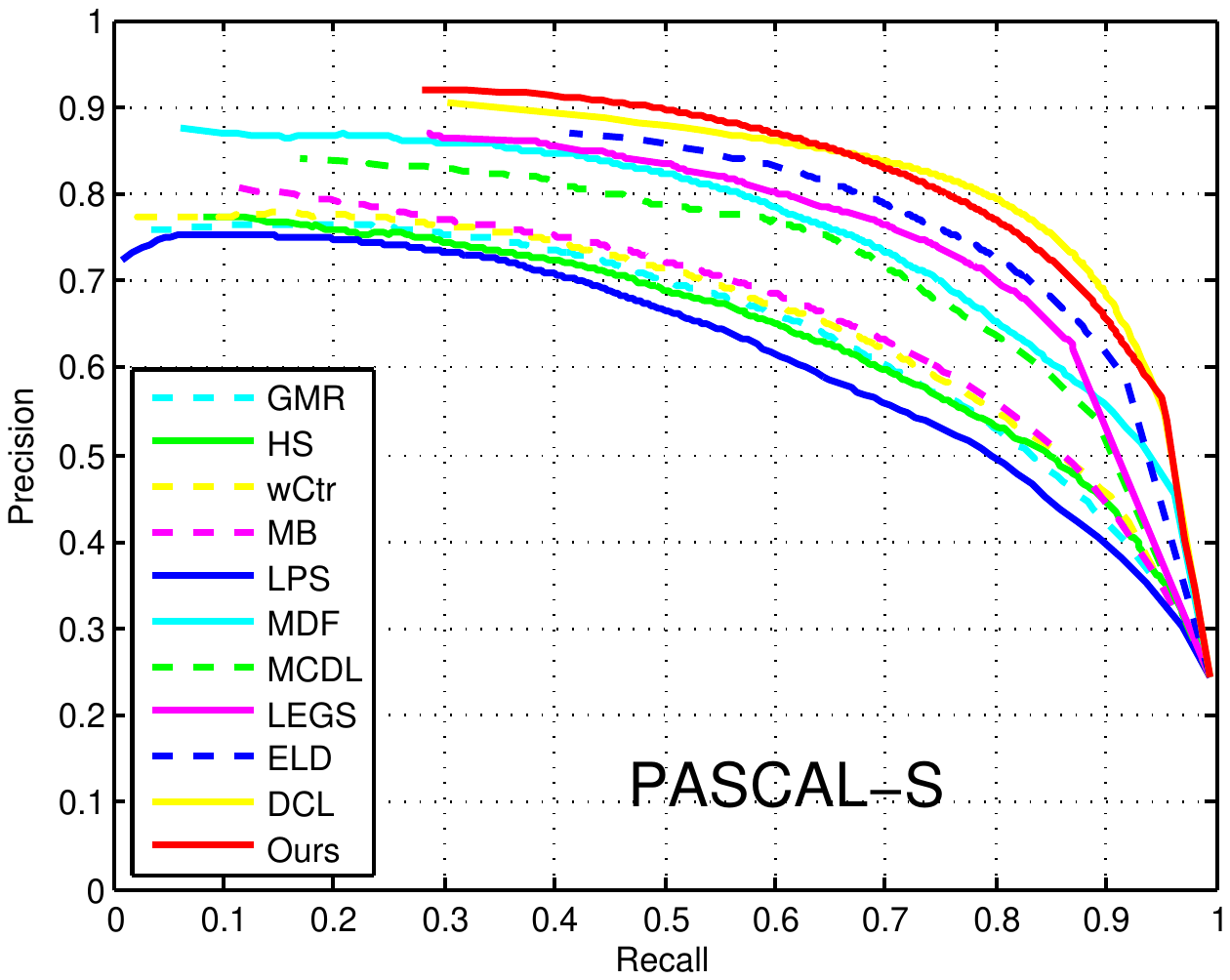}}
	\subfigure{\includegraphics[width=5.9cm]{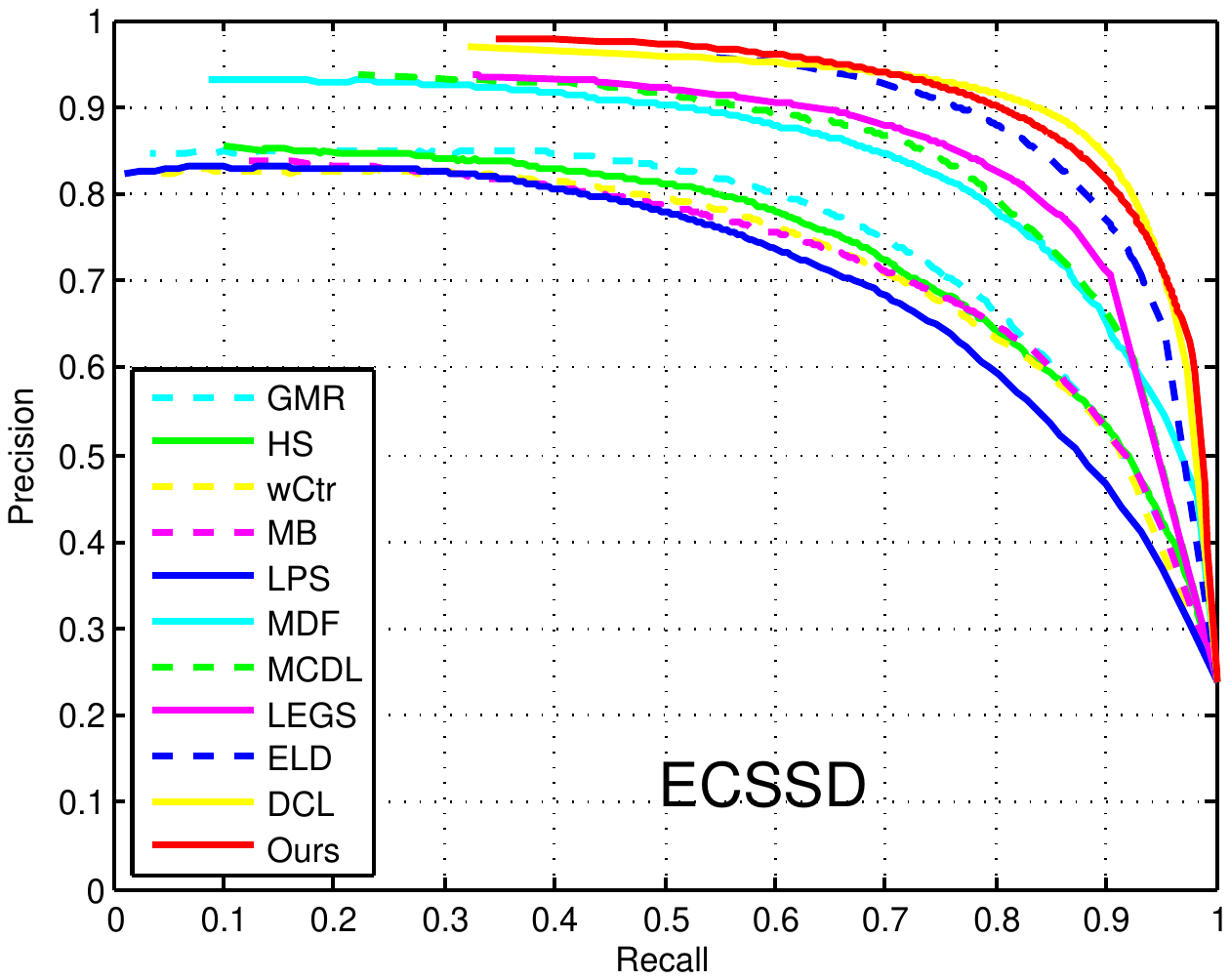}}
	\subfigure{\includegraphics[width=5.9cm]{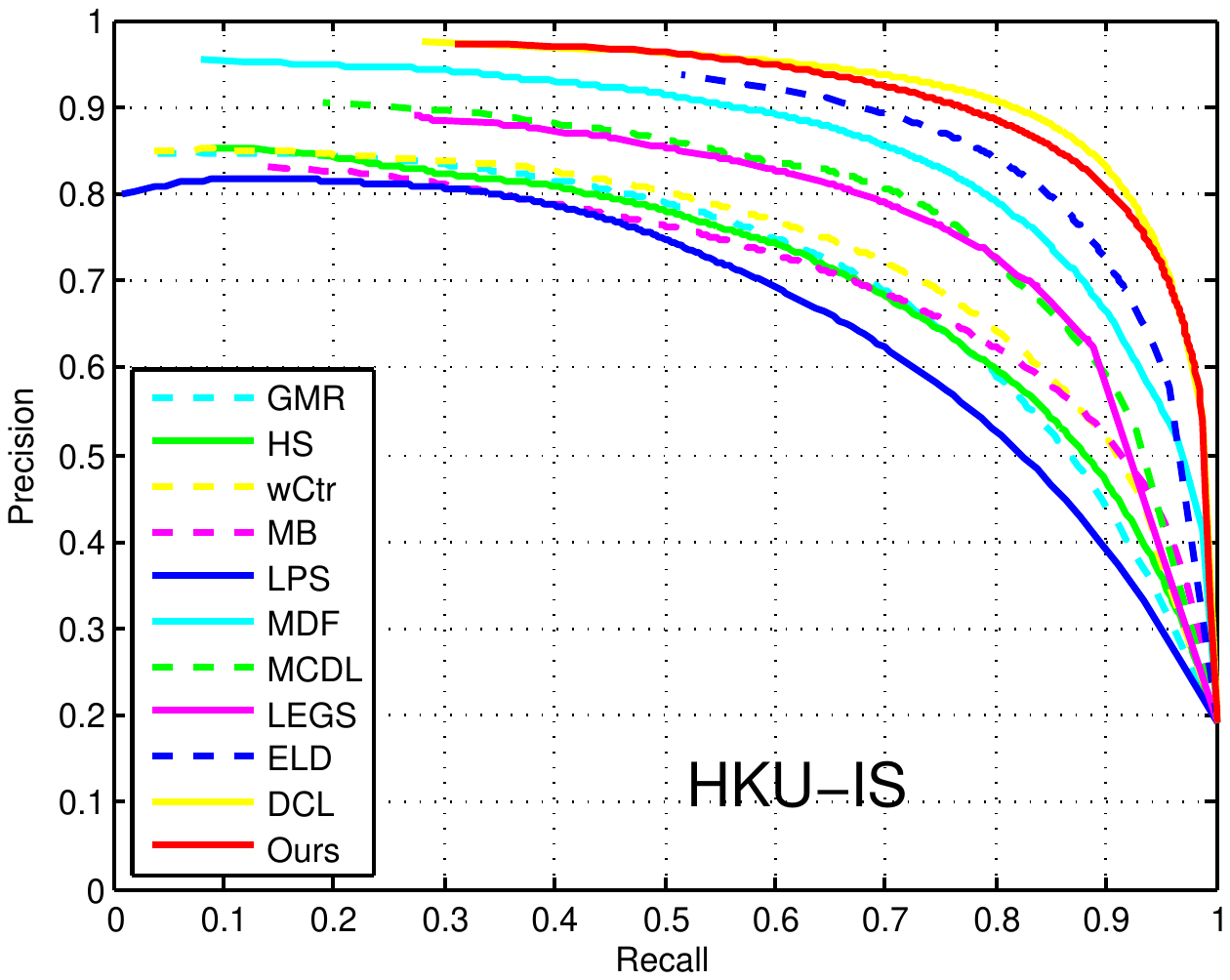}}
	\subfigure{\includegraphics[width=5.9cm]{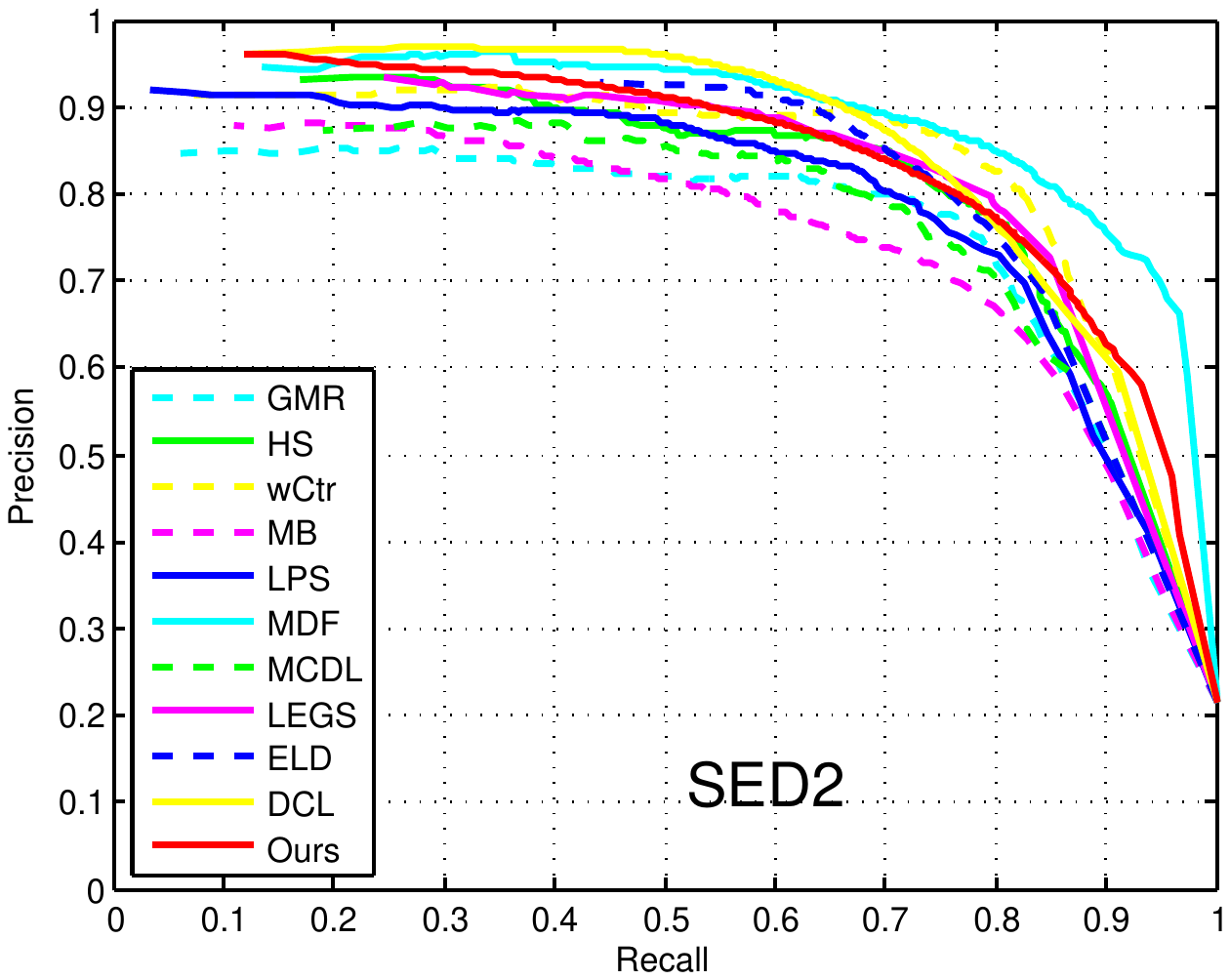}}
	\subfigure{\includegraphics[width=5.9cm]{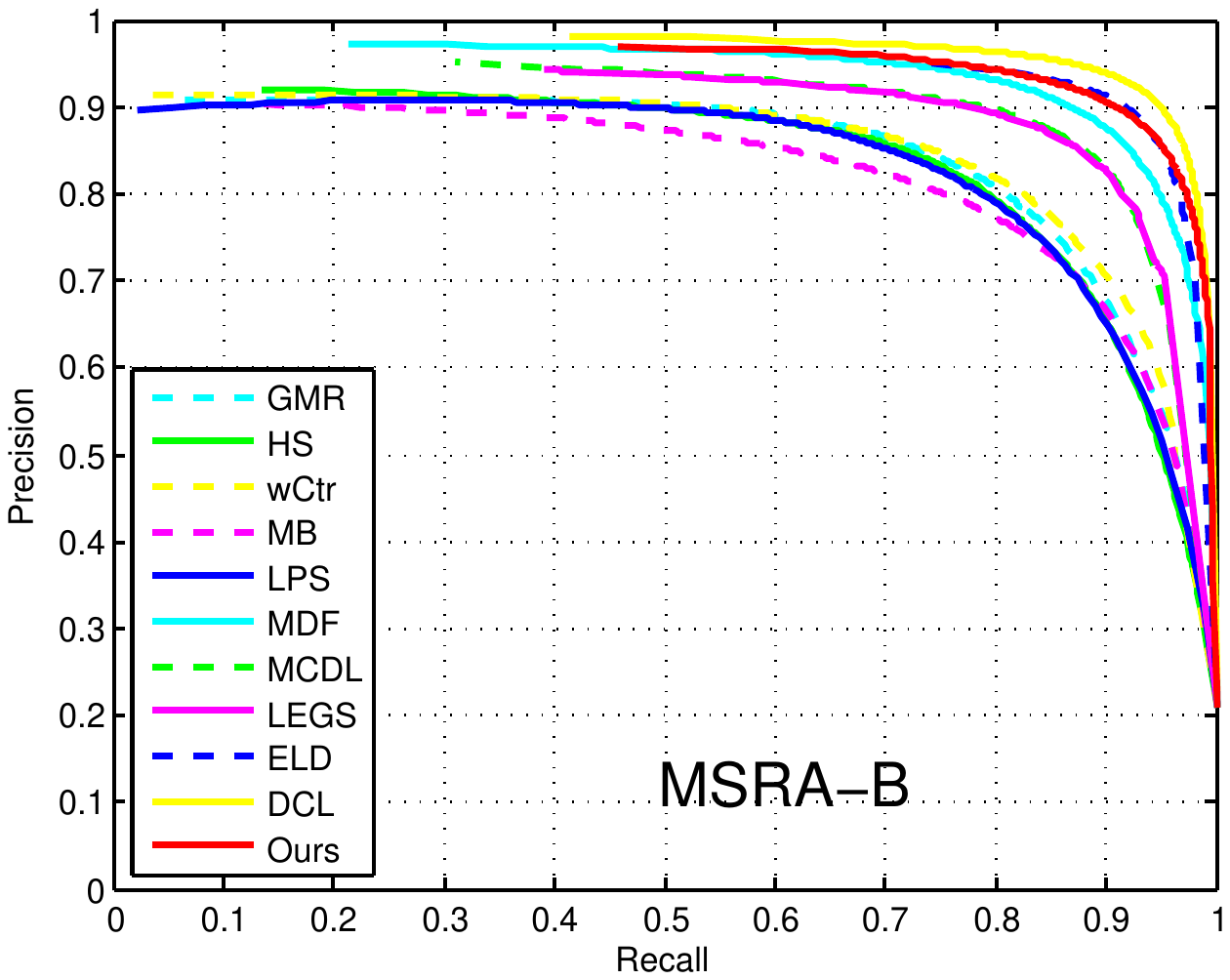}}
	\caption{PR curves of our method and 10 state-of-the-art methods on the 6 datasets.}
	\label{PRCurve}
\end{figure*}

\begin{figure*}[!t]
	\centering
	\subfigure{\includegraphics[width=5.9cm]{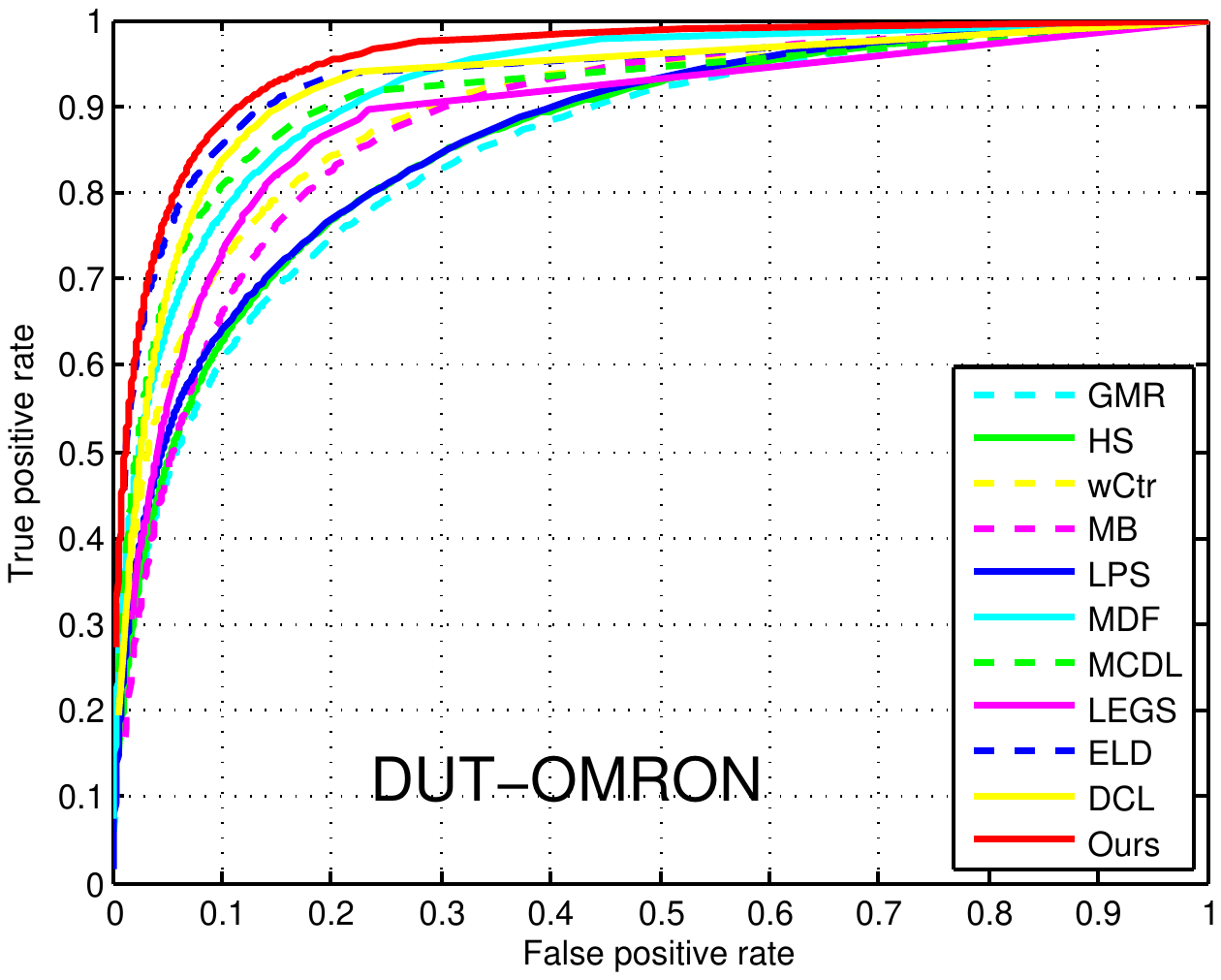}}
	\subfigure{\includegraphics[width=5.9cm]{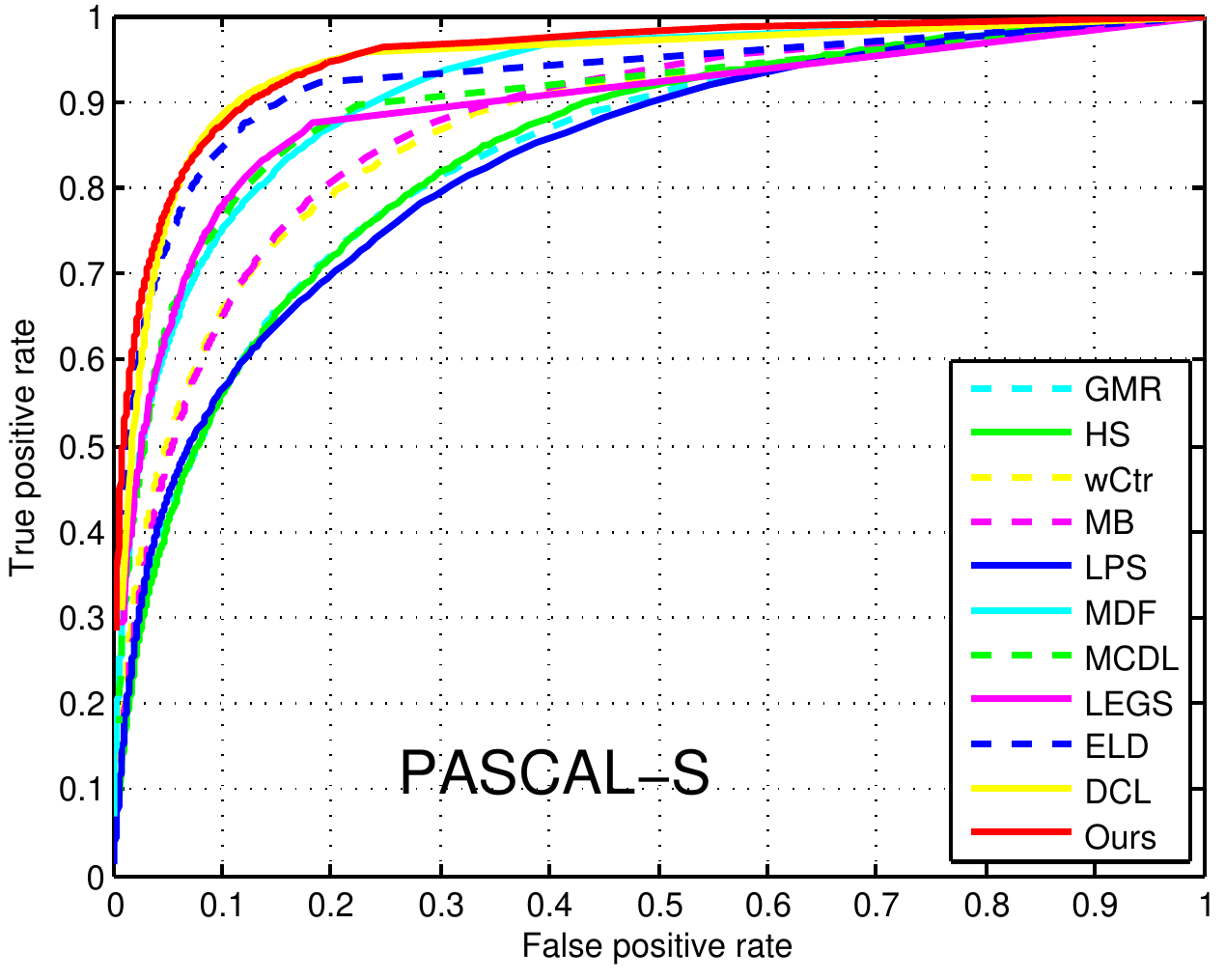}}
	\subfigure{\includegraphics[width=5.9cm]{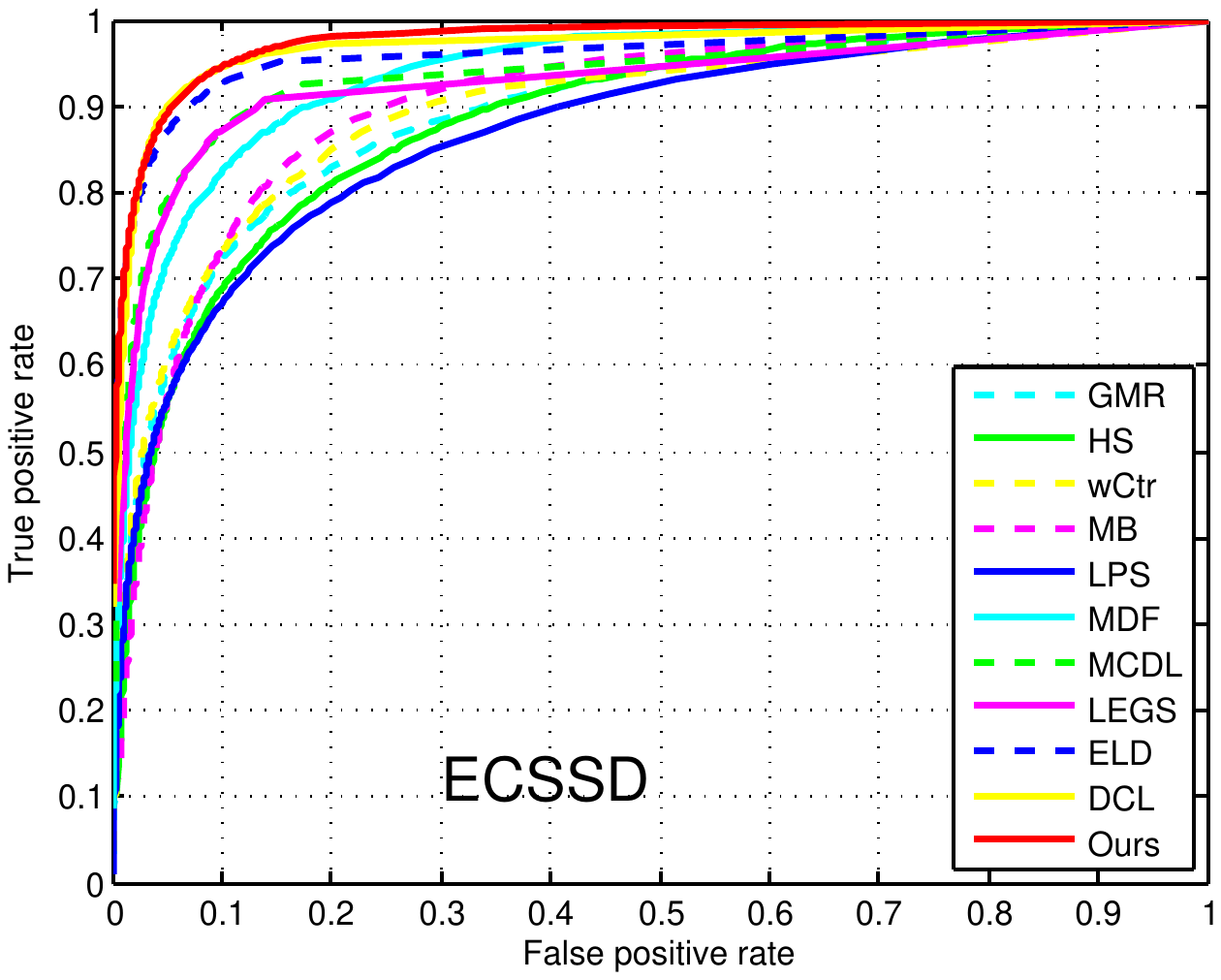}}
	\subfigure{\includegraphics[width=5.9cm]{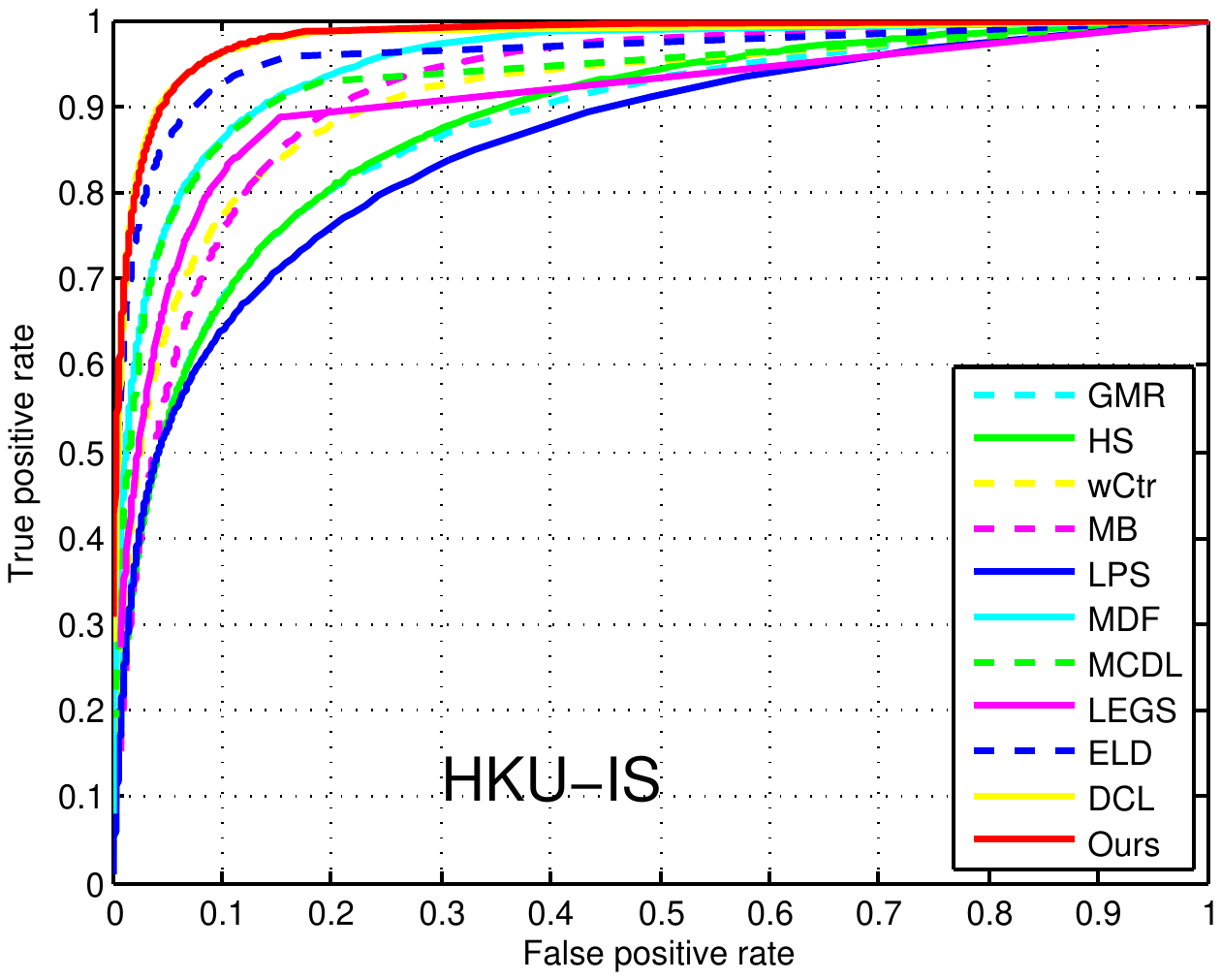}}
	\subfigure{\includegraphics[width=5.9cm]{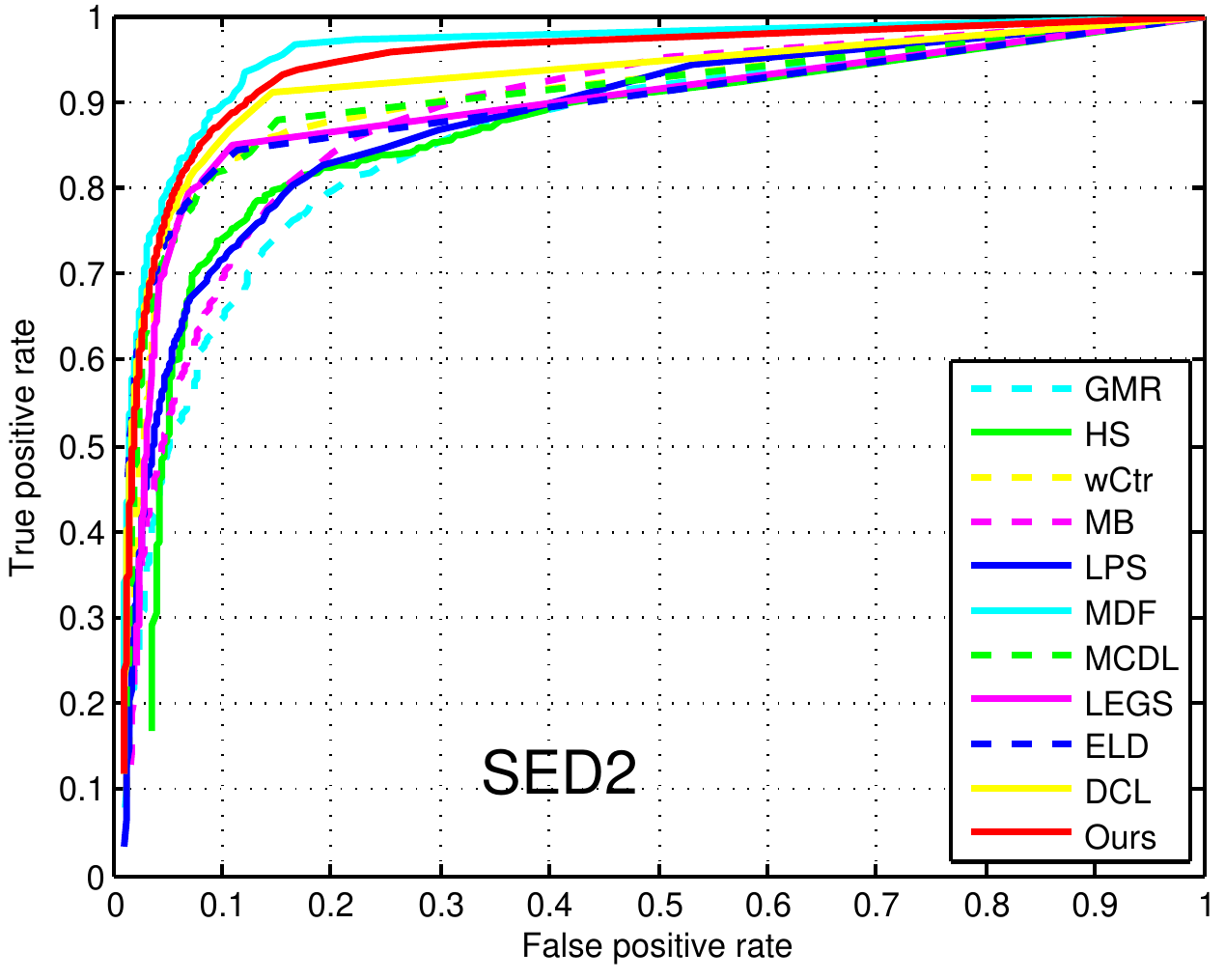}}
	\subfigure{\includegraphics[width=5.9cm]{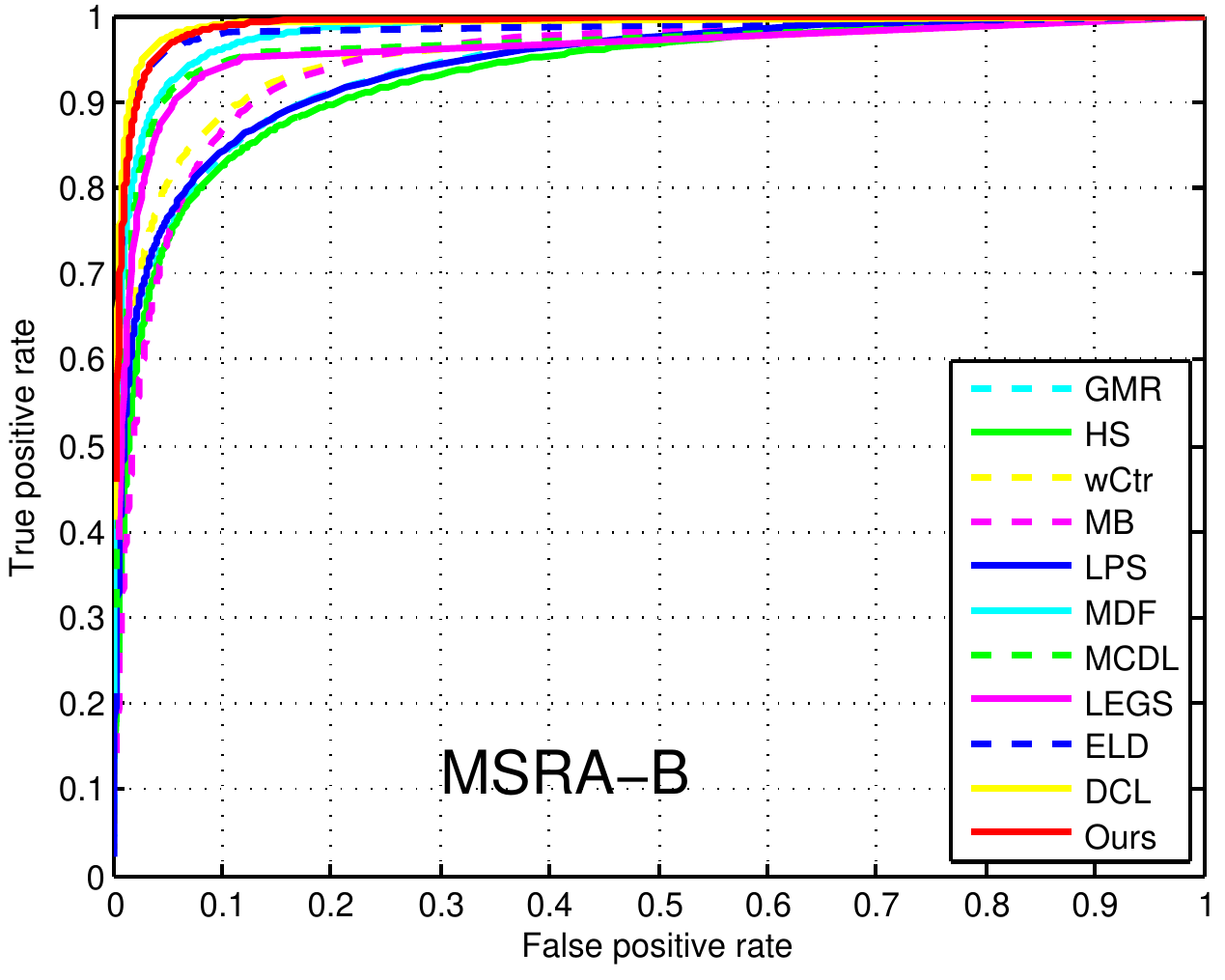}}
	\caption{ROC curves of our method and 10 state-of-the-art methods on the 6 datasets.}
	\label{ROCCurve}
\end{figure*}

Firstly, we plot the PR curve of all the methods on the 6 datasets in Fig.~\ref{PRCurve} and plot the ROC curve of all the methods on the 6 datasets in Fig.~\ref{ROCCurve}. From the two figures, we observe that our method achieves comparable or better performance than the other ones on most datasets. However, the PR curve of our method loses the advantage on the SED2 dataset. That is because we train our model on the MSRA10K whose images mostly contain single object while images in the SED2 have two salient objects. That is, the two datasets have different emphases which lead to the performance corruption. It is difficult to satisfy the uniqueness and completeness at the same time.

\begin{table*}[!t]
	\begin{center}
		\caption{ Comparison of quantitative results including AUC, F-meature (larger is better) and MAE, runtime (smaller is better). The best three results are respectively shown in \cR{red}, \cG{blue} and \cB{green} color. }
		\label{evaluationTable}
		\centering
		\scalebox{1}{
			\begin{tabular}{c | c | c c c c c | c c c c c c } 
				\hline \toprule
				Data        &Measure    &GMR  &HS    &wCtr  &MB    &LPS   &MDF   &MCDL  &LEGS  &ELD   &DCL   &Ours \\ \hline
				\multirow{4}{1.8cm}{DUT-OMRON}
				&AUC	    &0.852 &0.860 &0.894 &0.881 &0.866 &\cB{0.924} &0.912 &0.884 &\cG{0.932} &\cB{0.924} &\cR{0.954}\\
				&F-meature  &0.611 &0.616 &0.630 &0.589 &0.606 &0.680 &0.677 &0.668 &\cB{0.716} &\cG{0.717} &\cR{0.737} \\
				&MAE	    &0.187 &0.227 &0.144 &0.157 &0.145 &0.115 &\cG{0.094} &0.133 &\cR{0.092} &\cG{0.094} &\cB{0.100} \\
				&Runtime(s) &0.310 &0.322 &0.142 &\cR{0.010}&1.178 &13.30&1.625&1.214&0.575&\cB{0.135}&\cG{0.028}\\
				\hline				
				
				Data        &Measure    &GMR  &HS    &wCtr  &MB    &LPS   &MDF   &MCDL  &LEGS  &ELD   &DCL   &Ours \\ \hline
				\multirow{4}{1.8cm}{PASCAL-S}
				&AUC	    &0.836 &0.838 &0.866 &0.871 &0.830 &0.915 &0.897 &0.892 &\cB{0.924} &\cG{0.942} &\cR{0.950} \\
				&F-meature  &0.648 &0.641 &0.655 &0.663 &0.620 &0.733 &0.725 &0.749 &\cB{0.769} &\cR{0.803} &\cG{0.797} \\
				&MAE	    &0.231 &0.264 &0.201 &0.198 &0.219 &0.165 &0.148 &0.157 &\cG{0.123} &\cR{0.109} &\cB{0.128} \\
				&Runtime(s) &0.396&0.608&0.251&\cR{0.010}&1.426 &20.68&2.336&0.991&0.827&\cB{0.139}&\cG{0.042}\\
				\hline
				
				Data        &Measure    &GMR  &HS    &wCtr  &MB    &LPS   &MDF   &MCDL  &LEGS  &ELD   &DCL   &Ours \\ \hline
				\multirow{4}{1.8cm}{ECSSD}
				&AUC	    &0.891 &0.883 &0.894 &0.902 &0.871 &0.938 &0.934 &0.926 &\cB{0.957} &\cG{0.968} &\cR{0.976} \\
				&F-meature  &0.746 &0.730 &0.718 &0.714 &0.700 &0.807 &0.822 &0.831 &\cB{0.865} &\cR{0.886} &\cG{0.878} \\
				&MAE	    &0.188 &0.228 &0.173 &0.176 &0.188 &0.138 &0.107 &0.118 &\cG{0.081} &\cR{0.074} &\cB{0.087} \\
				&Runtime(s) &0.299&0.376&0.146&\cR{0.010}&1.102 &12.84&1.518&1.367&0.540&\cB{0.135}&\cG{0.027}\\
				\hline
				
				Data        &Measure    &GMR  &HS    &wCtr  &MB    &LPS   &MDF   &MCDL  &LEGS  &ELD   &DCL   &Ours \\ \hline
				\multirow{4}{1.8cm}{HKU-IS}
				&AUC	    &0.874 &0.879 &0.908 &0.916 &0.855 &0.951 &0.930 &0.905 &\cB{0.957} &\cG{0.977} &\cR{0.980} \\
				&F-meature  &0.709 &0.704 &0.723 &0.696 &0.673 &0.814 &0.780 &0.768 &\cB{0.841} &\cR{0.880} &\cG{0.866} \\
				&MAE	    &0.174 &0.215 &0.142 &0.149 &0.164 &0.112 &0.099 &0.119 &\cB{0.073} &\cR{0.058} &\cG{0.072} \\
				&Runtime(s) &0.307&0.381&0.140&\cR{0.009}&1.109 &12.17&1.518&1.442&0.537&\cB{0.136}&\cG{0.027}\\
				\hline	
				
				Data        &Measure    &GMR  &HS    &wCtr  &MB    &LPS   &MDF   &MCDL  &LEGS  &ELD   &DCL   &Ours \\ \hline
				\multirow{4}{1.8cm}{SED2}
				&AUC	    &0.856 &0.858 &0.899 &0.882 &0.876 &\cR{0.953} &0.901 &0.887 &0.892 &\cB{0.922} &\cG{0.941} \\
				&F-meature  &0.778 &0.811 &\cG{0.837} &0.733 &0.787 &\cR{0.843} &0.776 &0.811 &0.820 &\cB{0.831} &0.803 \\
				&MAE	    &0.164 &0.157 &0.130 &0.165 &0.141 &\cB{0.113} &0.120 &0.122 &\cG{0.104} &\cR{0.100} &0.114 \\
				&Runtime(s) &0.232&0.192&\cB{0.079}&\cR{0.009}&0.506 &4.825 &1.096&1.018&0.379&0.133&\cG{0.017}\\
				\hline	
				
				Data        &Measure    &GMR  &HS    &wCtr  &MB    &LPS   &MDF   &MCDL  &LEGS  &ELD   &DCL   &Ours \\ \hline
				\multirow{4}{1.8cm}{MSRA-B}
				&AUC	    &0.939 &0.930 &0.948 &0.945 &0.940 &\cB{0.981} &0.962 &0.958 &0.977 &\cR{0.990} &\cG{0.988} \\
				&F-meature  &0.821 &0.816 &0.823 &0.791 &0.811 &0.898 &0.872 &0.869 &\cG{0.912} &\cR{0.930} &\cB{0.910} \\
				&MAE	    &0.129 &0.162 &0.111 &0.124 &0.121 &0.072 &0.062 &0.082 &\cG{0.043} &\cR{0.040} &\cB{0.055} \\
				&Runtime(s) &0.317&0.324&0.146&\cR{0.009}&0.814 &11.72&1.569&1.394&0.567&\cB{0.135}&\cG{0.028}\\

				\bottomrule
				\hline
			\end{tabular}}
		\end{center}
	\vspace{-0.1in}
\end{table*}

In addition, we compare our method with the 10 methods in Table \ref{evaluationTable} with respect to AUC, F-meature, MAE and runtime. For a better view, we mark the best three results in each row with red, blue and green color respectively. As can be easily seen, the best detection precisions are all obtained by the deep learning based methods. Our method can be comparable with best approaches in terms of F-meature and MAE. What is more, our method can beat other methods in terms of AUC. In terms of runtime, the classical unsupervised methods are superior to deep learning based methods in general. Especially, the MB obtains the first place on all datasets. However, our method achieves the first place during above deep learning based methods with a big advantage (one order of magnitude or 4.8 times faster than the second fastest deep learning based method, DCL). Meanwhile, our method is also faster than other unsupervised methods except the MB. It is worth mentioning that the processing speed of our method can reach 35 FPS on the DUT-OMRON, ECSSD, HKU-IS, SED2 and MSRA-B. We find that the long side of images in these five datasets are less than 400 pixels. Then we can make a conclusion that our method can process images in 35 FPS, whose long sides are less than 400 pixels. It ensures that our method can run in a real time manner and can be practically used as a pro-processing step before other visual tasks.

In total, our method can be comparable with or better than recent state-of-the-art methods in precision, but makes a significant improvement in processing speed.

	
\subsection{Analyses of the Proposed Method}\label{AnalysisPart}
	
Through above comparison experiments, we present the characteristics of our method from various aspects. In this subsection, we plain to analyze some potential cues supporting the model's advantages through controlled experiments. We implement the controlled experiments only on large datasets (i.e. DUT-OMRON, HKU-IS and MSRA-B) which are more representative and impartial from a statistic point. Below we carry out the analyses from three aspects.
	
\subsubsection{Effectiveness of the Input Manner Based on Single Full-Resolution Images}\label{AnalysisA}
	
In the proposed method, we adopt the input manner based on single full-resolution images whose benefits we have explained in Section \ref{Architecture1}. To make the benefits more intuitionistic, we design a controlled experiment and name the corresponding model as controlled-model-A (CM-A for short hereafter). There is only one difference between CM-A and the proposed method that input images are resized to a uniform size ahead of being fed into CM-A, as done in \cite{li2015visual_MDF,ELD2016CVPR,DCL2016CVPR}. In detail, input images are resized to $224\times224$. As CM-A and the proposed method both have the same computational complexity, it is not required to compare their runtime. Their performance comparisons are summarized in Table \ref{Comparison-A}. It can be seen that the proposed method achieves better performance than CM-A on all items. Comparing with CM-A, the proposed method separately improves the AUC, F-meature and MAE performance by more than $1\%$, $3\%$ and $4\%$. The results can fully verify the effectiveness of our input manner.

\begin{table*}[!t]
	\caption{The peformance comparison between the proposed method and CM-A. The best results are shown in \cR{red} color.}
	\label{Comparison-A}
	\centering
	\begin{tabular}{p{2.8cm}<{\centering} | p{1.2cm}<{\centering} p{1.2cm}<{\centering} | p{1.2cm}<{\centering} p{1.2cm}<{\centering} | p{1.2cm}<{\centering} p{1.2cm}<{\centering} }
		\toprule
		\multirow{2}{2.4cm}{Data} &\multicolumn{2}{c|}{AUC} &\multicolumn{2}{c|}{F-meature} & \multicolumn{2}{c}{MAE} \\
		\cline{2-7}			
		            &Ours       &CM-A       &Ours       &CM-A       &Ours       &CM-A \\
		\hline
		DUT-OMRON   &\cR{0.954} &0.946      &\cR{0.737} &0.706      &\cR{0.100} &0.140 \\
		HKU-IS      &\cR{0.980} &0.968      &\cR{0.866} &0.828      &\cR{0.072} &0.115 \\					
		MSRA-B      &\cR{0.988} &0.976      &\cR{0.910} &0.869      &\cR{0.055} &0.112 \\
		\bottomrule
	\end{tabular}
\end{table*}

\subsubsection{Effectiveness of the Designed Loss Function}\label{AnalysisB}
	
In the proposed method, we design a suitable loss function to fit the characteristics of our network architecture. We have explained the design bases in Section \ref{lossdesign}. To make the design bases more intuitionistic, we design two controlled experiments and separately name corresponding models as controlled-model-B (CM-B for short) and controlled-model-C (CM-C for short). The two models only replace the loss function of the proposed method with other ones. The CM-B adopts classical Eudiance loss while the CM-C adopts cross entropy loss. Except this, the two model share the same input manner, network architecture, parameters and training strategy with the proposed method. Similarly, we compare the three models with respect to the three evaluation metrics and summary the results in Table \ref{Comparison-B}. The proposed method also achieves the best scores on all items. These two controlled experiments adequately verify the effectiveness of the designed loss function.

\begin{table*}[!t]
	\caption{The peformance comparison between the proposed method, CM-B and CM-C. The best results are shown in \cR{red} color. }
	\label{Comparison-B}
	\centering
	\begin{tabular}{p{2.6cm}<{\centering} | p{1cm}<{\centering} p{1cm}<{\centering} p{1cm}<{\centering} | p{1cm}<{\centering} p{1cm}<{\centering} p{1cm}<{\centering} | p{1cm}<{\centering} p{1cm}<{\centering} p{1cm}<{\centering} }
		\toprule
		\multirow{2}{2.6cm}{Data}  &\multicolumn{3}{c|}{AUC} &\multicolumn{3}{c|}{F-meature}& \multicolumn{3}{c}{MAE} \\ 
		\cline{2-4}	 \cline{5-7}	\cline{8-10}			
		&Ours       &CM-B  &CM-C       &Ours       &CM-B  &CM-C       &Ours       &CM-B  &CM-C \\
		\hline
		DUT-OMRON  &\cR{0.954} &0.945 &0.946      &\cR{0.737} &0.706 &0.718      &\cR{0.100} &0.124 &0.120 \\
		HKU-IS     &\cR{0.980} &0.978 &0.976      &\cR{0.866} &0.857 &0.861      &\cR{0.072} &0.092 &0.093 \\			
		MSRA-B     &\cR{0.988} &0.982 &0.982      &\cR{0.910} &0.880 &0.889      &\cR{0.055} &0.089 &0.088 \\
		\bottomrule
		
	\end{tabular}
\end{table*}

\subsubsection{Analysis of the Speed Improvement}\label{AnalysisC}

\begin{table*}[!t]
	\caption{Comparision of model size (the number of bytes required for storage). }
	\label{modelsize}
	\centering
	\begin{tabular}{ c| c c c c c c}
		\toprule
		Method	             &MDF     &MCDL           &LEGS        &ELD           &DCL    &Ours \\
		\hline
		Model Size (MByte)   &243.9   &(233.1+233.1) &(38.1+35.5) &(113.8+553.4) &265    &81.9 \\ 
		\bottomrule
	\end{tabular}
\end{table*}

In above two parts, we quantificationally analyze the effectiveness of the proposed method through controlled experiments. Further we try to analyze the potential cues supporting the speed improvement of the proposed method. It is meaningless to evaluate the speed performance separating with the precision performance. Then it is cumbersome to orderly analyze corresponding factors through controlled experiments. Here we just give a brief analysis intuitively.

From Table \ref{evaluationTable}, we can find that our method makes a significant improvement in processing speed compared with other deep learning based methods. After carefully pondering the network architecture, we think that four cues contribute to the success: a) we adopt an image-level saliency score regression framework not a salient region classification framework as in \cite{ELD2016CVPR,MCDL2015CVPR}; b) our method does not rely on superpixel-level or region-level segmentation maps as in \cite{ELD2016CVPR,MCDL2015CVPR,li2015visual_MDF}, which at lest take 400 ms; c) our method does not utilize time-consuming post-processing approaches like CRF \cite{DCL2016CVPR}; d) we simplify the network architecture and abandon some tricks which can improve the precision performance, such as multi-scale \cite{li2015visual_MDF,DCL2016CVPR} and combing local and global information \cite{wang2015deep_LEGS,MCDL2015CVPR}.

In fact, these cues can also reduce the number of learned parameters and shorten the required training time effectively. We summarize the model size (the number of bytes required for storage) of aforementioned deep learning based methods in Table \ref{modelsize}. As the model size is directly proportional to the accurate number of learned parameters, we employ the model size to make an illustration. In general, our method owns a smaller storage compared with others. Although the model size of LEGS is a little smaller than ours (around 8 MByte), its performances including both precision and runtime are clearly inferior to ours (refering to Table \ref{evaluationTable}). The small parameter storage also increases the usability of our method.

\section{Conclusion}
	
In this paper, we propose a fast and compact saliency score regression network to meet the essential application requirement of salient object detection task. In the architecture aspect, it is constructed by modifying the VGG16-Net and adopts a single full-resolution input manner. In the loss function aspect, we design a suitable loss function to meet the characteristic of the proposed network architecture. There is no additional pre-processing and post-processing beyond the network. After training the network can directly predict a dense full-resolution saliency map when being fed into an image. The end-to-end work manner effectively simplifies the processing procedure. By comprehensive experiments, we verify that our method can achieve comparable or better precision performance than the state-of-the-art methods while get a significant improvement in detection speed (processing in real time). The compact architecture, fast processing speed, small parameter storage and decent precision performance make it possible to employ our method practically as a pre-processing step before other visual tasks. 

In the future work, we try to further perfect the model by adding proper post-processing approaches. Now the proposed method can process images in 35 FPS. It still has some space to improve its precision performance but within the scope of processing in real time.

%

%

{
\small
\bibliographystyle{IEEEtran}
\bibliography{references}
}

\end{document}